\documentclass[ijoc,nonblindrev]{informs3} 

\OneAndAHalfSpacedXI 

\usepackage{musicography}
\usepackage{natbib}
\usepackage{mathtools}
\usepackage{algorithm}
\usepackage{algpseudocode}
\usepackage{booktabs}
\usepackage{multirow}

\usepackage{pgfplots} 
\usepackage{subcaption}
\usepackage{caption}
\usetikzlibrary{arrows.meta}

\usepackage{color,soul}
 \bibpunct[, ]{(}{)}{,}{a}{}{,}%
 \def\bibfont{\small}%
\TheoremsNumberedThrough     
\ECRepeatTheorems

\algnewcommand{\Inputs}[1]{%
  \State \textbf{Inputs:}
  \Statex \hspace*{\algorithmicindent}\parbox[t]{.8\linewidth}{\raggedright #1}
}
\algnewcommand{\Outputs}[1]{%
  \State \textbf{Outputs:}
  \Statex \hspace*{\algorithmicindent}\parbox[t]{.8\linewidth}{\raggedright #1}
}
\algnewcommand{\Initialize}[1]{%
  \State \textbf{Initialize:}
  \Statex \hspace*{\algorithmicindent}\parbox[t]{.8\linewidth}{\raggedright #1}
}

\EquationsNumberedThrough    

\MANUSCRIPTNO{JOC-2022-05-OA-136.R1}
\usepackage{caption}
\usepackage[hidelinks]{hyperref}
\usetikzlibrary{patterns}
\def\eg{\emph{e.g.}}
\def\ie{\emph{i.e.}}
\begin{document}


\RUNAUTHOR{Böttcher, Asikis, and Fragkos}

\RUNTITLE{Control of Dual-Sourcing Inventory Systems using Recurrent Neural Networks}

\TITLE{Control of Dual-Sourcing Inventory Systems using Recurrent Neural Networks}

\ARTICLEAUTHORS{%
\AUTHOR{Lucas Böttcher}
\AFF{Dept.~of Computational Science and Philosophy, Frankfurt School of Finance and Management, 60322 Frankfurt, Germany, \EMAIL{l.boettcher@fs.de}}
\AFF{Dept.~of Computational Medicine, University of California, Los Angeles, Los Angeles, CA 90095-1766, USA}
\AUTHOR{Thomas Asikis}
\AFF{Game Theory, University of Zurich, 8092 Zurich, Switzerland, \EMAIL{thomas.asikis@uzh.ch}}
\AUTHOR{Ioannis Fragkos}
\AFF{Dept.~of Technology and Operations Management, Rotterdam School of Management, Erasmus University Rotterdam, 3062 Rotterdam, Netherlands}
}

\ABSTRACT{%
A key challenge in inventory management is to identify policies that optimally replenish inventory from multiple suppliers. To solve such optimization problems, inventory managers need to decide what quantities to order from each supplier, given the net inventory and outstanding orders, so that the expected backlogging, holding, and sourcing costs are jointly minimized. Inventory management problems have been studied extensively for over 60 years, and yet even basic dual-sourcing problems, in which orders from an expensive supplier arrive faster than orders from a regular supplier, remain intractable in their general form. In addition, there is an emerging need to develop proactive, scalable optimization algorithms that can adjust their recommendations to dynamic demand shifts in a timely fashion. In this work, we approach dual sourcing from a neural network\textendash based optimization lens and incorporate information on inventory dynamics and its replenishment (\ie, control) policies into the design of recurrent neural networks. We show that the proposed neural network controllers (NNCs) are able to learn near-optimal policies of commonly used instances within a few minutes of CPU time on a regular personal computer. To demonstrate the versatility of NNCs, we also show that they can control inventory dynamics with empirical, non-stationary demand distributions that are challenging to tackle effectively using alternative, state-of-the-art approaches. Our work shows that high-quality solutions of complex inventory management problems with non-stationary demand can be obtained with deep neural-network optimization approaches that directly account for inventory dynamics in their optimization
process. As such, our research opens up new ways of efficiently managing complex, high-dimensional inventory dynamics.
}%


 \HISTORY{This paper is in preprint stage.}

\maketitle

%


\section{Introduction}
Inventory management problems of various forms have been studied for more than six decades by the operations management and operations research communities. Progress in solving such problems has led to effective and efficient supply chains of unprecedented size and complexity. Despite some celebrated results, such as the optimality of base-stock policies in single-sourcing systems with backlogging~\citep{scarf1958inventory}, most inventory management problems have such complex dynamics that finding optimal policies has been a major challenge. For example, in inventory management problems with two suppliers (\ie, dual-sourcing problems), the general structure of the optimal policy remains unknown after more than half a century of intense research~\citep{sun2019robust, goldberg2021survey}. Since dual sourcing and other inventory management problems are analytically intractable, a vast amount of effort has been put into designing heuristics that guide effective order decisions. 

In this work, we study dual sourcing in its classic form, as first analyzed by \cite{barankin1963delivery} and \cite{fukuda1964optimal}, from a neural-network optimization lens. We develop inventory management systems where order policies are represented by neural networks and we show that the proposed optimization method is able to approximate the structure of optimal policies and outperform effective dual-sourcing heuristics in commonly used instances. We also show that the proposed neural-network optimization methods are able to effectively control complex inventory dynamics with demand distributions that are inferred from empirical data.

Using state-of-the-art machine learning methods combined with domain-specific knowledge is a relatively new and promising venue for inventory management research \citep{xin2021dual,song2020capacity}. An important contribution in this direction is the work by \cite{gijsbrechts2020can}, who utilize deep reinforcement learning (RL) to control three archetypal inventory management problems. The authors show that an actor-critic RL algorithm can deliver competitive performance against certain problem-specific heuristics. This is an important step towards the design of more generic algorithms that are able to solve a larger set of optimization problems with less restrictive assumptions. In this paper, we contribute to this line of research, and more generally to solving discrete stochastic optimization problems, by introducing neural network controllers (NNCs) that directly control dual-sourcing dynamics by learning effective order policies without relying on RL methods. 

The optimization methods that we develop in this article build on previous works that use automatic differentiation~\citep{linnainmaa1976taylor,paszke2017automatic,baydin2018automatic} and dynamics-informed neural networks~\citep{raissi2019physics} to solve complex control and optimization problems~\citep{DBLP:conf/iclr/HollTK20,DBLP:conf/nips/JinWYM20,DBLP:journals/corr/abs-2003-00868,asikis2020nnc,bottcher2021implicit}. We construct recurrent neural networks that take both the current inventory and previous orders as inputs and minimize the expected backlogging, holding, and sourcing costs. There are two major challenges that arise in training neural networks for controlling inventory dynamics. First, the evolution of inventory systems is described by a stochastic dynamical system, so neural networks have to be trained on a sufficiently large number of realizations to learn effective and generalizable order policies.\footnote{In the context of controlling stochastic dynamical systems, we refer to a neural network\textendash based control policy as ``generalizable'' if the learned policy, trained on specific realizations of the stochastic dynamics, is able to achieve similar performance (\eg, a similar loss) on unseen realizations.} In order to design networks that learn generalizable order policies, we utilize activation functions that resemble the structure of optimal solutions of simple inventory models, namely basestock policies. Second, adjusting neural-network weights during training relies on propagating real-valued gradients, while neural-network outputs (\ie, replenishment orders) are required to be integers. To solve such a discrete optimization problem with real-valued gradient-descent learning algorithms, we employ a straight-through estimator technique~\citep{asikis2021multi,yang2022injecting} to obtain integer-valued neural-network outputs while backpropagating real-valued gradients.
\subsection{Contributions}
\label{sec:contributions}
In summary, our work contributes to the extant inventory management literature in the following aspects.
\begin{itemize}
    \item We show that the proposed NNCs are able to approximate optimal inventory management policies without any previous knowledge on their structure. In addition to comparing optimal objective values with the corresponding values obtained by neural networks, we compare, whenever possible, neural-network policies with the optimal ones. This form of comparison, although uncommon in the literature, sheds light on the structure of replenishment policies and their resemblance to optimal ones. Consequently, we show evidence that neural networks are able to uncover policies that resemble the structure of optimal policies, instead of simply attaining near-optimal objective values with policies of arbitrary structure. Solutions found by the proposed NNCs can, in some cases, be interpreted as generalizations of state-of-the-art heuristic policies~\citep{jiang2019service}.
    \item Our approach is computationally attractive since it requires only between a few seconds and a few minutes of CPU time on a regular personal computer. To the best of our knowledge, this is the first generic approach that can compete and even outperform tailored policies using a reasonable amount of computational power. A key element in the algorithm that we use to effectively learn replenishment policies with neural networks is a problem-tailored straight-through estimator~\citep{bengio2013estimating,yin2019understanding} that allows a neural network to output discrete actions (\ie, integer-valued quantities) and still adjust real-valued neural-network weights. Using a straight-through estimator enables us to embed inventory-state calculations into NNCs and define their recurrent connections in a model-based manner, introducing an inductive bias that can improve learning performance~\citep{baxter2000model}.
    \item Finally, we demonstrate the versatility of NNCs by using them to effectively control instances of dual-sourcing problems with fixed order costs~\citep{svoboda2021typology} and with empirically inferred, non-stationary demand distributions~\citep{manary2021data}. In the majority of simulated realizations with empirical demand data, we find that NNCs outperform a state-of-the-art inventory control heuristic. For dual-sourcing problems with fixed order cost, the optimal policy structure is not known and no good heuristics are available~\citep{svoboda2021typology}. Still, NNC order policies are associated with expected ordering costs that are close to the optimal ones, which we approximate with a dynamic-programming approach.
\end{itemize}

The remainder of this paper is organized as follows. In Section~\ref{sec:literature}, we review previous work on inventory management problems, and we discuss similarities and differences between the proposed NNCs and related neural network\textendash based control and optimization methods. Section~\ref{sec:nn_optimization} formulates the generic problem of controlling discrete-time stochastic dynamical systems with recurrent neural networks and it discusses how tailored straight-through estimators can be incorporated into a backpropagation algorithm to effectively learn near-optimal inventory replenishment policies. In Section~\ref{sec:example}, we illustrate the methodology on a simple inventory model with a single supplier, synthetic demand data and backlogging, establishing a correspondence between the network architecture and the structure of the optimal solution of the mode. We then customize NNCs to dual-sourcing problems in Section~\ref{sec:nnc_inventory} and present computational experiments in a wide variety of instances in Section~\ref{sec:computations}. In the first part Section~\ref{sec:computations}, we focus on synthetic demand profiles with and without fixed order costs while the last part of Section~\ref{sec:computations} applies neural-network order policies to dual-sourcing problems with empirical demand data. Finally, we conclude our work in Section~\ref{sec:conclusion}, reflecting on avenues for future research.
\section{Literature Review}
\label{sec:literature}
We next review some relevant work from the inventory management literature and the rapidly growing field of neural network\textendash based control and optimization.
\subsection{Inventory Management with Dual Sourcing}
\label{sec:dual_sourcing_heuristics}

Research in dual-sourcing inventory management problems is vast and spans over more than five decades. In what follows, we overview seminal papers, papers that investigate the structure of optimal solutions under special cases, papers that focus on developing effective heuristics, and recent literature reviews. For the sake of brevity, we primarily focus on related work that studies dual-sourcing dynamics in its original form.

The core problem where one can order from a cheaper but slower and a faster but more costly supplier using a dynamic order allocation rule was first studied by \cite{barankin1963delivery} and \cite{fukuda1964optimal}, who showed that a single-index, dual-base-stock policy is optimal when the supplier lead times are one time period apart. When lead times differ more than one period, the optimal policy is state-dependent, as it depends on the vector of pipeline orders in a non-trivial way \citep{whittemore1977optimal}. As the difference between the supplier lead times grows larger, the problem becomes computationally intractable because of the associated ``curse of dimensionality'' \citep{powell2007approximate, goldberg2021survey}. Most of the literature has therefore focused on the development of heuristic policies, such as the single index (SI) \citep{scheller2007effective}, dual index (DI) \citep{veeraraghavan2008now}, capped dual index (CDI) \citep{sun2019robust}, vector-base-stock (VBS) \citep{sheopuri2010new, hua2015structural}, and tailored base-surge (TBS) \citep{allon2010global, xin2018asymptotic, chen2019tailored} policies. 

Although the structure of optimal solutions remains unknown, some interesting properties have been derived. First, \cite{sheopuri2010new} observe that an inventory dynamics model with both a single supplier and lost sales can be derived as a special case of a dual-sourcing system in which the fast supplier is able to deliver an order instantaneously, \emph{after} the period's demand has been realized. For appropriately chosen parameters, it is optimal to order from that supplier exactly as many items are necessary to clear the backlog, and therefore the corresponding ordering cost can be seen as a lost sales cost in a lost-sales model. The same authors observe that the state space can be compressed to $l$ dimensions, where $l$ is the lead time difference between the two suppliers, and that, without loss of generality, the expedited supplier's lead time can be set to zero when working in the compressed space. Second, following the analysis of \cite{zipkin2008structure}, \cite{hua2015structural} characterize the structure of optimal solutions by showing that the value function is $L^{\musNatural}$ convex. In particular, they show that regular orders are more sensitive to late-arriving orders than to earlier ones, while expedited orders are more sensitive to the net inventory position. Their result is used to construct a heuristic policy that performs very well against other approaches. In a subsequent study, \cite{xin2018asymptotic} show that TBS is asymptotically optimal as the lead time of the regular supplier grows large and the lead time of the fast supplier remains fixed. Tailored base surge is also near-optimal when demand comprises two components, a base distribution and a surge distribution, provided that the surge demand occurs with a small probability \citep{janakiraman2015analysis}. Finally, \cite{sun2019robust} show that CDI policies are robustly optimal, given demand that lies in a known polyhedral uncertainty set. The authors also establish that in the limiting case where the regular supplier's lead time grows large and the lead time of the expedited supplier remains unchanged, CDI converges to TBS, thereby matching the main result of \cite{xin2018asymptotic}. Capped dual index policies achieve a good performance over a wide range of lead time differences and resemble DI policies for small lead time differences \citep{xin2021dual}, presumably because the effect of an additional parameter that describes order constraints (or ``caps'') is negligible. Capped base-stock policies also exhibit good performance on single-sourcing, lost-sales inventory models \citep{xin20211, xin2021understanding}. Adding a cap to a base-stock policy reduces the variance of ordered quantities, which may be advantageous in lost-sales models. Intuitively, while in backlogging systems an excessively large order can help clearing a large backlog, in lost-sales models it will only increase the inventory on hand and possibly the holding costs. This property of lost-sales models, namely the period ``reset'' of inventory to zero whenever demand is in excess, makes near-myopic and capping policies effective in practice \citep{morton1971near, zipkin2008old, xin2021understanding}.

Dual sourcing can be extended in multiple other directions, such as having endogenous stochastic lead times \citep{song2017optimal}, multi-source systems \citep{song2021smart}, non-stationary demand, and supplier capacities \citep{boute2021dual}. A comprehensive overview of dual-sourcing models is presented in \cite{xin2021dual}, while \cite{goldberg2021survey} provide a review on asymptotic analysis in inventory control problems and show how it can be applied to dual-sourcing models. Finally, a very comprehensive overview of the recent literature can be found in \cite{svoboda2021typology}, where the authors provide a typology of various inventory control models with multiple suppliers, stages, time periods, and uncertainty structures.

Our paper augments the above literature by offering an alternative perspective on identifying effective order policies in dual-sourcing inventory systems. To the best of our knowledge, ours is the first approach that shows how neural networks can compete with and even outperform inventory-management approaches developed after years of specialized research on dual sourcing. From a methodological perspective, our paper is close to \cite{gijsbrechts2020can}, who apply RL-based policy learning to a variety of inventory management problems. In particular, the authors show that the presented Asynchronous Advantage Actor-Critic (A3C) algorithm competes favorably with certain heuristics for lost-sales, dual-sourcing, and multi-echelon models. The authors note that the initial tuning phase can be computationally burdensome. Evaluating one set of neural-network hyperparameters may take about 24 CPU hours. We build upon this research by using NNCs that control the stochastic difference equations underlying dual-sourcing dynamics. When tested on datasets from the literature, our approach outperforms state-of-the-art dual-sourcing heuristics, using a few seconds up to a few minutes of CPU time on a regular personal computer. In addition, our approach outperforms competitive approaches when tested on empirical demand data \citep{manary2021data}. Using data-driven machine learning models for inventory control is a burgeoning area of research that can address practical inventory management problems \citep{svoboda2021typology}. Indeed, our results provide evidence that the presented neural network control methods are effective for solving standard and non-standard dual-sourcing problems.
\subsection{Neural Network\textendash based Optimization and Control}
\label{sec:nn_optimization_control}
The NNCs that we develop in this work to manage inventory replenishment have their origin in control theory and machine learning. In the context of inventory management, optimal control signals correspond to the orders that minimize the expected shortage, holding, and sourcing costs, given the system state (\ie, net inventory and outstanding orders). Such control approaches are connected to recent advances in automatic differentiation and physics-informed neural networks~\citep{karniadakis2021physics,raissi2019physics,lutter2019deep,zhong2019symplectic,mowlavi2023optimal}, which have found applications in modeling partially unknown systems~\citep{roehrl2020modeling}. In physics-informed neural networks, one constrains the training process by including information on the spatio-temporal evolution of a physical system into the loss function. Such constraints are typically described by partial differential equations. In our work, the loss function is proportional to the expected total cost (\ie, the expected sum of backlogging, holding, and sourcing costs over a certain time horizon) and we incorporate additional model-specific information in the training process by using the state variables of inventory dynamics (\ie, net inventory and outstanding orders) as inputs.

For deterministic, continuous-time dynamical systems with real-valued control signals, control frameworks that are based on neural ordinary differential equations (ODEs) were proposed by \cite{asikis2020nnc}. In a related work by \cite{bottcher2021implicit} and \cite{bottcher2022near}, it has been shown that such neural ODE controllers are able to learn control trajectories that resemble those of optimal control methods. Another work by \cite{asikis2021multi} extends the aforementioned framework with real-valued control signals to discrete action spaces. Here, we build on this discrete action-space formulation and study the ability of neural networks to control inventory dynamics with \emph{stochastic} demand.

Designing neural network\textendash based control methods for discrete action spaces is challenging since standard neural-network optimization techniques rely on learning suitable \emph{real-valued weights} based on the propagation of \emph{real-valued gradients}. \cite{chen2020learning} use neural event functions to model discrete and instantaneous changes in a continuous-time system. The work of \cite{asikis2021multi} proposes fractional decoupling, a straight-through estimator~\citep{bengio2013estimating,yin2019understanding} to perform real-valued gradient calculations for control problems with discrete action spaces. Fractional decoupling calculates gradients based on real-valued variables, instead of using discrete values that are applied in the cost-function calculation of control problems with discrete action spaces. In this way, it is possible to use standard neural network optimization techniques and still generate discrete control signals (\eg, discrete order quantities in inventory management problems).

To explain the main difference between NNCs and RL-based optimization approaches, we briefly summarize their respective design features and application areas. In RL, an \emph{agent} performs a certain \emph{action} in a surrounding \emph{environment}~\citep{sutton2018reinforcement}. Actions are mapped to a \emph{reward}, which is used to determine an optimal action sequence for a given sequence of states. The ultimate goal is to solve an underlying optimization problem (\ie, the Bellman equation) iteratively using a Markov decision process. There are two major classes of RL algorithms: (i) model-free and (ii) model-based methods. The main difference between these two classes is that the former uses a trial-and-error approach in determining optimal actions, while the latter uses a model to describe interactions between agent and environment~\citep{polydoros2017survey}. Model-free methods are widely used, but may converge slowly and entail the risk of exploring unfavorable actions~\citep{yarats2021improving}. If an appropriate model can be used to explore the action space, model-based approaches may be a favorable alternative since they converge faster towards the desired solution that maximizes the reward function. Still, model-based approaches are often based on neural networks representing policy and value functions, and require a differentiable model that fully describes the environment, which might be challenging to derive for high-dimensional and complex control tasks~\citep{Wang2018}.

If the deterministic or stochastic dynamics that describe a certain system are known, an alternative to RL-based approaches is to directly represent the policy function by a single neural network and backpropagate gradients resulting from a gradient descent in a problem-specific loss function. For inventory management problems, such as dual sourcing, the underlying dynamics can be formulated in terms of stochastic difference equations, and an NNC can be constructed to learn the action in a particular state that minimizes the expected backlogging, holding, and sourcing costs.

The neural network control approaches that we use in this work have several advantages over model-free RL methods, which suffer from sample inefficiency resulting from underlying high-dimensional feature spaces~\citep{yarats2021improving,jin2018q}. Reinforcement-learning algorithms are often based on learning target and behavior policies. The target policy is the policy or neural network that updates its own parameters. The behavior policy is the policy that samples trajectories and selects actions based on state observations.

In on-policy learning, the value function is learned from actions that are based on the current policy. In this case, behavior and target policies are the same. In off-policy learning, the value function is learned using different (\eg, random) actions. Advantage Actor-Critic and other actor-critic algorithms are primarily on-policy while progress in deep RL also led to the development of effective off-policy actor-critic approaches~\citep{DBLP:conf/icml/DegrisWS12,DBLP:conf/icml/SchmittHS20} that have been used to improve sample efficiency~\citep{DBLP:conf/iclr/0001BHMMKF17}. However, off-policy algorithms are vulnerable to three error accumulation processes: (i) erroneous approximation of the target function (function approximation error), (ii) divergence of the target policy from the behavior policy (divergence error), and (iii) calculation of a value estimate by the target policy for an action-state pair based on value estimates of other action pairs (bootstrapping error). These three sources of error are known as the ``deadly triad'' of RL~\citep{van2018deep,sutton2018reinforcement}. By not using a value function\textendash based RL framework, we avoid diverging and bootstrapping errors, two main challenges associated with the deadly triad. Since the neural network control methods that we develop in this work are model-based and on-policy, they do not explicitly rely on value-function approximations obtained through randomly sampling trajectories and they do not explicitly use value estimates to derive their policy. In addition, the policies that NNCs learn are used to generate future samples for learning, so they do not suffer from the off-policy divergence problem.

To summarize, neural network control approaches provide an effective tool to represent and learn complex policy functions associated with known deterministic and stochastic dynamical systems~\citep{asikis2020nnc,bottcher2021implicit}. Deep reinforcement learning is preferable in scenarios where the interactions between agent and environment cannot be fully characterized in terms of a mechanistic model.
\section{Control of Discrete-Time Stochastic Dynamical Systems}
\label{sec:nn_optimization}
As a starting point, we consider a discrete-time stochastic dynamical system
\begin{equation}
\mathbf{s}_{t+1}=\mathbf{f}(\mathbf{s}_t,\mathbf{a}_t,D_t)\,,
\label{eq:discrete_time_evolution}
\end{equation}
where $\mathbf{s}_t$ and $\mathbf{a}_t$ denote state and action variables at time $t$, respectively. $D_t \sim \phi$ is a random variable, and $\mathbf{f}$ maps the current state, action, and realization of $D_t$ to a new state $\mathbf{s}_{t+1}$. In the inventory dynamics systems that we study in this work, actions $\mathbf{a}_t$ represent integer-valued replenishment orders and depend only on the system state $\mathbf{s}_t$ [\ie, $\mathbf{a}_t=\mathbf{a}_t(\mathbf{s}_t)$] for stationary demand distributions $\phi$. Each state-action combination is associated with a cost $c_t(\mathbf{s}_t,\mathbf{a}_t)$. Our goal is to identify a policy $\pi_t=\{\mathbf{a}_{t+j}|j=0,1,\dots\}$ (\ie, a sequence of actions) that minimizes the expected total cost over $T$ periods,
\begin{equation}
C_t^{(\pi_t)}(\mathbf{s}_t,T)=\sum_{j=0}^T \gamma^j \mathbb{E}^{(\pi_t)}\left[c_{t+j}(\mathbf{s}_{t+j},\mathbf{a}_{t+j})\right]\,,
\label{eq:cost_function_1}
\end{equation}
where $\gamma \in (0,1]$ is a discount factor and $\mathbb{E}^{({\pi}_t)}\left[c_{t+j}\right]$ is the expected cost at time $t+j$ that results from policy ${\pi}_t$. 

The infinite-time mean expected cost of a policy $\pi_t$, which is independent of the initial state $\mathbf{s}_t$, is defined as

\begin{equation}
    \label{eq:average_exp_cost}    
    J^{(\pi_t)} = \lim_{T \rightarrow \infty}\sup \left\{ \frac{1}{T}  C_t^{(\pi_t)}(\mathbf{s}_t, T)\right\}\,.
\end{equation}

The objective of the discrete stochastic optimization problems that we study in this work is to identify the optimal policy $\pi^*$ that yields the minimum expected cost per period over an infinite horizon, \ie, $\pi^* \in \arg \inf_{\pi_t} J^{(\pi_t)}$.
\subsection{Neural Network\textendash based Policies}
Instead of pursuing a dynamic-programming approach, we parameterize actions using neural networks with parameters $\mathbf{w}$. For a stationary demand distribution, we denote the corresponding neural-network policies and actions by $\hat{\pi}_t=\{\hat{\mathbf{a}}_{t+j}|j=0,1,\dots\}$ and $\hat{\mathbf{a}}_t=\hat{\mathbf{a}}_t(\mathbf{s}_t;\mathbf{w})$, respectively. Note that an optimal, non-randomized policy associated with minimizing Eq.~\eqref{eq:average_exp_cost} maps each state-space vector $\mathbf{s}_t$ to a corresponding action vector $\mathbf{a}_t$. The focal problem of the neural-network control approach that we pursue in this work is therefore to determine the function that maps states to actions such that Eq.~\eqref{eq:average_exp_cost} is optimized. Theoretically, neural networks are able to represent such policies under mild conditions, as specified by universal approximation theorems \citep{hornik1991approximation,hanin2017approximating,park2020minimum}. In practice, however, designing and training neural networks that learn optimal policies of discrete-time stochastic dynamical systems has been challenging~\citep{gijsbrechts2020can,boute2021deep}. We next outline an algorithm that does so efficiently.

Using minibatches of size $M$, we first approximate $\mathbb{E}^{(\pi_t)}\left[c_{t+j}(\mathbf{s}_{t+j},\mathbf{a}_{t+j})\right]$ in Eq.~\eqref{eq:cost_function_1} by
\begin{equation}
\mathbb{E}^{(\hat{\pi}_t)}\left[c_{t+j}(\mathbf{s}_{t+j},\hat{\mathbf{a}}_{t+j})\right] = \frac{1}{M}\sum_{k=1}^M c_{t+j}(\mathbf{s}^{(k)}_{t+j},\hat{\mathbf{a}}^{(k)}_{t+j})\,,
\label{eq:expected_cost_minibatch}
\end{equation}
where $\mathbf{s}^{(k)}_{t}$ and $\hat{\mathbf{a}}^{(k)}_{t}$ denote state and neural-network actions in minibatch $k$ at time $t$. This empirical approximation of $\mathbb{E}^{(\pi_t)}$ is then used in conjunction with Eq.~\eqref{eq:cost_function_1} to obtain
\begin{align}
\hat{J}^{(\hat{\pi}_t)} = \frac{1}{T} \sum_{j=0}^T \gamma^j \frac{1}{M} \sum_{k=1}^M c_{t+j}(\mathbf{s}_{t+j}^{(k)},\hat{\mathbf{a}}_{t+j}^{(k)})\,,
\label{eq:emp_loss}
\end{align}
the empirical long-time mean expected cost associated with neural network policy $\hat{\pi}_t$.
\begin{figure}
    \FIGURE
    {\includegraphics[width=\textwidth]{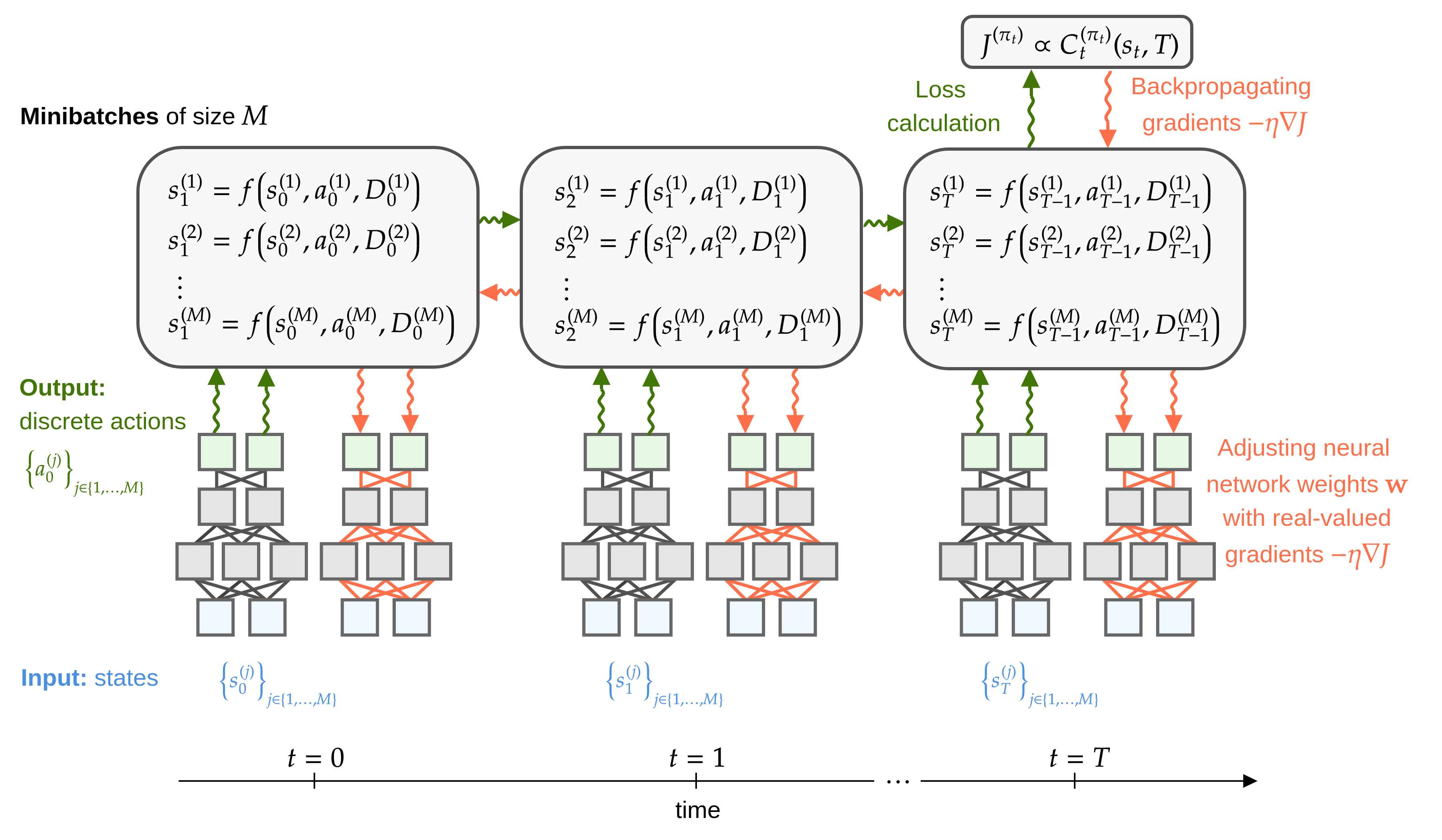}}
    {Schematic of solving discrete-time stochastic control problems with neural networks.\label{fig:optimization_schematic}}
    {}
\end{figure}
Equation \eqref{eq:emp_loss} shows that losses are accumulated over time. 

To learn effective policy functions, we compute the final value of the loss function $\hat{J}^{(\hat{\pi}_t)}$ for a given time horizon $T$ in each training epoch and then adjust neural network parameters $\mathbf{w}$ using backpropagation through time (BPTT), a common training algorithm for recurrent neural networks (RNNs)~\citep{williams1990efficient,werbos1990backpropagation,feldkamp1993neural}. 
The corresponding gradient-descent updates of neural-network weights $\mathbf{w}$ are given by
\begin{align}
\label{eq:backprop}
\mathbf{w}^{(n+1)}=\mathbf{w}^{(n)}-\eta \nabla_{\mathbf{w}^{(n)}} 
\hat{J}^{(\hat{\pi}_t)}\,,
\end{align}
where $n$ and $\eta$ denote the current training epoch and learning rate, respectively. In our numerical experiments, we primarily use the adaptive learning rate algorithm RMSProp~\citep{rmsprob} to perform gradient updates (see the e-companion for further details).

A schematic of the described optimization process is shown in Figure~\ref{fig:optimization_schematic}. States $\{\mathbf{s}_t^{(j)}\}$ ($1\leq j \leq M$) that evolve according to the discrete-time dynamics \eqref{eq:discrete_time_evolution} are used as inputs in a neural network that learns actions that minimize the loss function $\hat{J}^{(\hat{\pi}_t)}$. Similar to ODE-Nets and neural ODEs~\citep{wang1998runge,DBLP:conf/nips/ChenRBD18}, minimizing the loss function $\hat{J}^{(\hat{\pi}_t)}$ is associated with a time-unfolded neural network structure that can be seen as an RNN because an action $\hat{\mathbf{a}}_t(\mathbf{s}_t;\mathbf{w})$ (\ie, the neural network output) at period $t$ affects the state $\mathbf{s}_{t+1}$ (\ie, the input of the same neural network) at the subsequent period $t+1$.

Two key challenges in training RNNs with BPTT are exploding gradients and vanishing gradients. Another limitation of BPTT is that its runtime increases with the number of periods $T$. In certain applications, it may not be feasible to construct a computational graph for backpropagation if the number of periods is too large. There exist different numerical recipes that can be used to address these issues. For example, gradient-clipping techniques are useful to avoid exploding gradients while training RNNs with BPTT. In addition, truncated versions of the standard BPTT algorithm can help avoid vanishing gradients and runtime issues. The basic idea of truncated BPTT (TBPTT) is to detach certain parts in the loss function $J^{(\hat{\pi}_t)}$ from the computational graph (\eg, costs, states, and actions for periods smaller than a certain threshold) that is used to perform gradient updates. However, a disadvantage of TBPTT protocols is that they discard information (\eg, certain state-demand combinations) that might be important to learn the right actions. Because the neural networks that we employed in this work were able to learn near-optimal policies with a good runtime performance and without vanishing gradient problems, we decided not to discard certain parts of the loss function. We also performed BPTT updates without gradient clipping as we did not encounter exploding gradients during training.
\subsection{Straight-Through Estimators and Fractional Decoupling}
\label{sec:frac_decoupling}
To apply the described neural network\textendash based policies to inventory management problems, the outputs of the underlying neural network $\{\hat{\mathbf{a}}_t^{(j)}\}$ ($1\leq j \leq M$) are required to be discrete and non-negative as order quantities are typically integer-valued. However, standard backpropagation algorithms assume that neural-network parameters and outputs are real-valued~\citep{rumelhart1986learning,lecun2012efficient}. Different techniques have been proposed in the literature to overcome this issue, including (i) rounding neural-network outputs by employing a linear relaxation principle~\citep{lin2019approximate} or a nearest-neighbor learning approach~\citep{dulac2015deep}, (ii) assigning a real-valued score/probability to each potential action (\eg, Q-values~\citep{sutton2018reinforcement} or Temporal Difference Error~\citep{ieee_rl_discrete}) to then select actions with high scores~\citep{gijsbrechts2020can}, and (iii) applying neural search approaches such as neural branch-and-bound algorithms~\citep{nair2020solving}.

Rounding is non-differentiable and thus cannot be directly combined with backpropagation training. Furthermore, the optimal real-valued outputs of a trained neural network may not round to the optimal integer solution. Using algorithms to learn effective rounding protocols~\citep{dulac2015deep} introduces additional computational costs. In RL, learning (discrete) actions requires one to learn a corresponding value function, which is often performed in a model-free manner. As pointed out in Section~\ref{sec:nn_optimization_control}, model-free RL approaches may suffer from different issues as value estimation may be sample inefficient~\citep{botvinick2019reinforcement} and, depending on the choice of the RL algorithm, convergence may become difficult to achieve~\citep{van2018deep}. When the state transition probabilities are known, one can derive a model-based dynamic programming approach to efficiently estimate values associated with corresponding actions, which, however, may also be computationally expensive. Finally, a combination of dynamic-programming approaches and neural network\textendash based optimization methods~\citep{nair2020solving} may also incur high computational costs.

The straight-through estimation technique~\citep{bengio2013estimating,yin2019understanding} that we use in conjunction with RNNs to learn effective inventory management policies differs in several aspects from the optimization methods mentioned above. In a straight-through estimator, a certain mathematical operation that is applied in a forward pass is treated as an identity operation in the backward pass (\ie, when calculating gradients using backpropagation). The straight-through estimator that we implement in this work rounds neural-network outputs by subtracting from the positive parts $[\mathbf{y}^{(j)}]^+$ of the outputs $\mathbf{y}^{(j)}$ ($1\leq j \leq M$) of the last hidden layer (\ie, the layer before the output layer) their corresponding fractional parts $\{[\mathbf{y}^{(j)}]^+\}$. That is, in a forward pass, the neural network output is
\begin{equation}
\hat{\mathbf{a}}^{(j)}_t = [\mathbf{y}^{(j)}]^+ - \{[\mathbf{y}^{(j)}]^+\}\,,
\label{eq:fract_decoupling}
\end{equation}
where $\{x\}=x - \lfloor x \rfloor$ if $x>0$ and $\lfloor \cdot \rfloor$ denotes the floor function. While updating neural-network weights by backpropagating gradients, we detach the fractional part $\{[\mathbf{y}^{(j)}]^+\}$ from the underlying computational graph (\ie, the neural-network is treated as $\hat{\mathbf{a}}^{(j)}_t = [\mathbf{y}^{(j)}]^+$). We therefore refer to this problem-tailored straight-through estimator as \emph{fractional decoupling}. 

In contrast to the aforementioned model-free techniques, the use of fractional decoupling allows a neural network to output discrete values and enables us to directly incorporate inventory dynamics and the corresponding cost calculation in the RNN optimization as shown in Figure \ref{fig:optimization_schematic}. Effectively, we update neural-network weights according to Eq.~\eqref{eq:backprop} and compute gradients that contain information on inventory dynamics by using automatic differentiation and straight-through estimators. The proposed model-based approach is an alternative to model-based value estimators, such as dynamic programming, which use state transition probabilities and value estimates to select optimal actions.

To summarize, including both inventory dynamics and rounding capabilities in an NNC structure during training provides a possibility to facilitate faster convergence by introducing a form of inductive bias~\citep{baxter2000model}. In the remainder of this work, we first discuss an example related to single-sourcing dynamics to motivate the use of the outlined NNCs in inventory management. We then provide experimental results on controlling dual-sourcing dynamics that show that the proposed control method is able to learn near-optimal replenishment policies and outperform state-of-the-art heuristics.
\section{Motivating Example}
\label{sec:example}
Modern neural networks utilize several layers of activation functions to represent high-dimensional functions. One of the most commonly used activation function is the Rectified Linear Unit (ReLU), which returns the positive part of a real number, \ie, ${\rm ReLU}(x)=\max\{0, x\}$~\citep{DBLP:conf/icml/NairH10,schmidhuber2015deep}. This is, in fact, equivalent to the structure of base-stock policies, a class of policies that have been shown to be optimal for a variety of inventory control problems, including single-source problems with backlogged demand \citep{scarf1958inventory}, batch ordering \citep{veinott1965optimal}, multiple products under resource constraints \citep{decroix1998optimal}, and Markov-modulated demand \citep{song1993inventory}. To illustrate how this connection can help NNCs to effectively replenish inventory, we first focus on inventory management problems with a single supplier, whose optimal solutions are known to have a basestock structure. For such problems, a single ${\rm ReLU}$ neuron is able to represent the optimal structure, and it motivates the use of ${\rm ReLU}$ activations and its variants in more complicated problems, such as dual sourcing. Although such learning is theoretically possible, it is not trivial that it is computationally efficient. Indeed, without fractional decoupling training becomes so slow that it is unlikely to lead to competitive performance when generalized to dual-sourcing problems. We describe the single sourcing model next.
\subsection{Single Sourcing Model and Optimal Policy}
To mathematically describe the optimal order policy of single sourcing problems~\citep{arrow1951optimal,scarf1958inventory}, we use $l$ and $z$ to respectively denote the replenishment lead time and the target inventory-position level (\ie, the target net inventory level plus all goods on order). The inventory position of single-sourcing dynamics at time $t$, $\tilde{I}_t$, is given by
\begin{equation}
\tilde{I}_t=
\begin{cases}
I_t\quad &\text{if} \,\, l=0\\
I_t+\sum_{i=1}^l q_{t-i} \quad &\text{if} \,\, l>0\,,
\end{cases}
\end{equation}
where $I_t$ and $q_t$ denote the net inventory at time $t$ and the replenishment order placed at time $t$, respectively. We let $b$ and $h$ denote the unit backlogging and holding costs, respectively. The optimal target inventory level~\citep{arrow1951optimal} is given by the critical fractile
\begin{equation}
z^* = \Phi^{-1}\left(\frac{b}{b+h}\right)\,,
\end{equation}
where $\Phi(x)=\Pr(D\leq x)$ denotes the cumulative distribution function of demand $D$ during $l+1$ periods. If the inventory position falls below $z^*$ at time $t$, a replenishment order $q_t=z^*-\tilde{I}_t$ is placed to bring the inventory position back to the optimal target level. The optimal single-sourcing policy (or ``base stock'') is thus
\begin{equation}
q_t=[z^*-\tilde{I}_t]^+\,.
\label{eq:optimal_base_stock}
\end{equation}

We observe that the optimal single-sourcing order policy is given by the positive part of $z^*-\tilde{I}_t$, which depends on the optimal inventory-position level $z^*$, the current net inventory, and the sum of previous orders $q_{t-i}$ ($1 \leq i\leq l$). 
\subsection{Designing a Neural Network Controller}
To construct an NNC that learns replenishment orders $\hat{q}_t$, we use $l+1$ inputs that represent the known net inventory and previous orders (\ie, the system state). We also include a bias term in the input layer to model the unknown optimal target inventory level $z^*$. All inputs are passed into an activation function that generalizes expression Eq.~\eqref{eq:optimal_base_stock}. Using a ReLU activation function will match exactly the structure of the optimal policy.
However, while updating the weights of a ReLU activation using \eqref{eq:backprop}, it may end up in an inactive state in which it produces near-zero outputs (\eg, due to a large negative bias term)~\citep{douglas2018relu}. Once a ReLU reached such a state, it is unlikely to recover because corresponding gradients almost vanish and gradient-descent learning will not be able to substantially alter the weights; hence, the output will stay close to zero. We avoid this so-called ``dead ReLU'' problem by using a continuously differentiable exponential linear unit
\begin{equation}
{\rm CELU}(x,\alpha)=\left[x\right]^+ - \left[\alpha \left(1-\exp(x/\alpha)\right)\right]^+\,,
\label{eq:celu}
\end{equation}
which approaches ${\rm ReLU}=[x]^+$ in the limit $\alpha\rightarrow 0^+$~\citep{barron2017continuously}. An advantage of CELUs over ReLUs is that they are continuously differentiable, facilitating the gradient calculation in neural-network parameter optimization.

\begin{figure}
    \FIGURE
    {\includegraphics{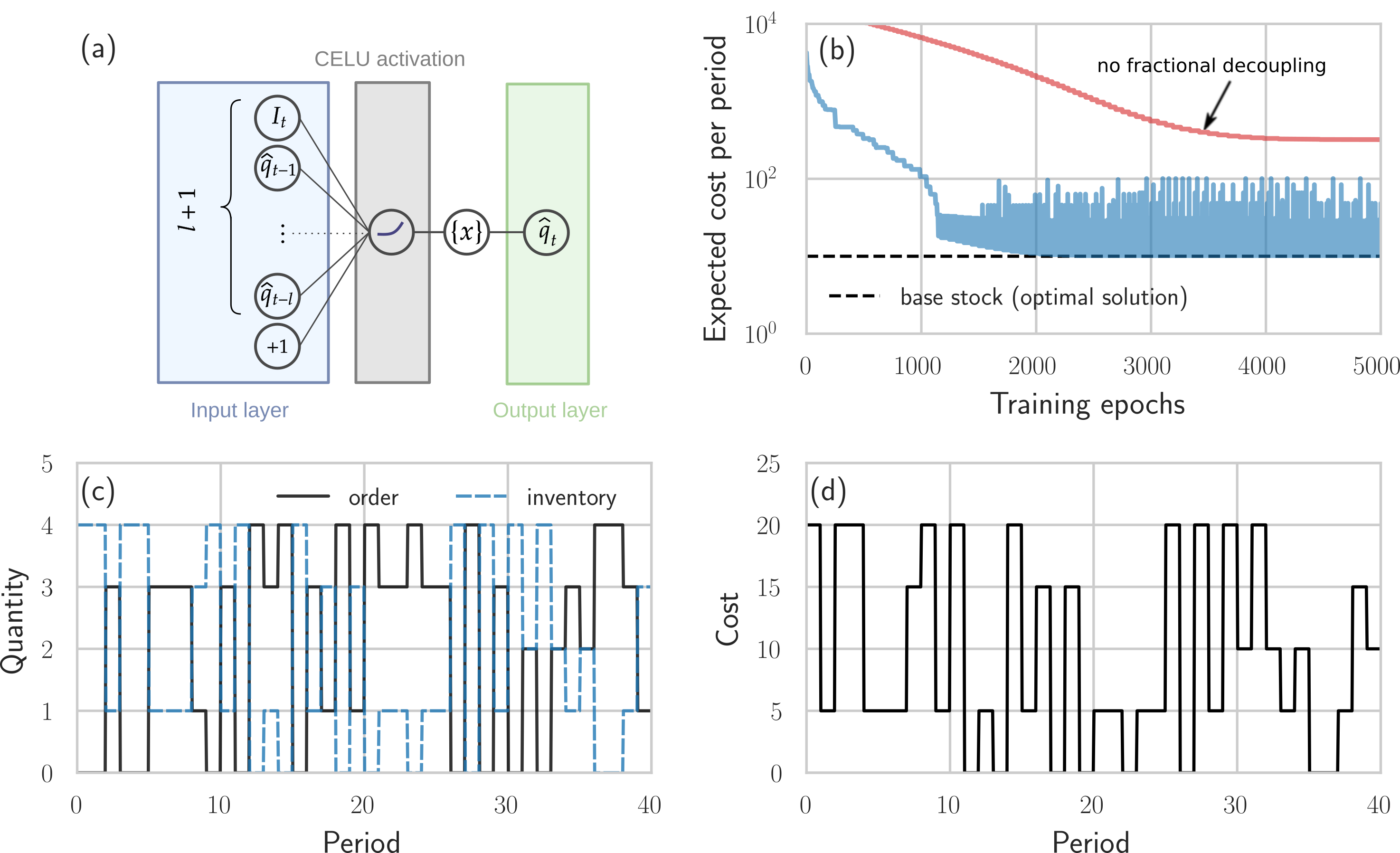}}
    {Controlling single-sourcing problems with neural networks.\label{fig:base_stock}}
    {(a) Schematic of an NNC architecture for single-sourcing dynamics. The neural network uses as input the ${(l+1)}$-dimensional state $\mathbf{s}_t=[I_{t},\hat{q}_{t-1},\dots,\hat{q}_{t-l}]^\top$ and outputs a non-negative integer-valued action $\hat{a}_t\equiv \hat{q}_t$ after subtracting the fractional part $\{[y_2]^+\}$ of the positive part of the CELU output $[y_2]^+$ [see Eq.~\eqref{eq:fract_decoupling}]. In addition to $\mathbf{s}_t$, the input layer also includes a bias term, indicated by a ``+1''. The hidden layer is composed of one CELU activation. (b) Expected cost per period as a function of training epochs for $h=5$, $b=495$, $l=0$, $T=50$, and demand distribution $\mathcal{U}\{0,4\}$. The solid red line shows the expected cost evolution for training with a rounding layer and no fractional decoupling (no convergence towards the optimal solution). In all computations, we set $\alpha=1$ in the CELU activation. (c) Evolution of inventory and orders for an optimal base stock level of four units. (d) The corresponding cost evolution.}
\end{figure}

Figure~\ref{fig:base_stock}(a) shows a schematic of the neural-network architecture that we use to learn the optimal single-sourcing policy.
We denote the input vector (\ie, the system state) and the output of the linear input layer by $\mathbf{s}_t=[I_t,\hat{q}_{t-1},\dots,\hat{q}_{t-l}]^\top\in\mathbb{R}^{l+1}$ and $y_1=\mathbf{w}_1^\top \mathbf{s}_t+e_1$, respectively. The CELU activation function in the second layer mimics the $[\cdot]^+$ operation in Eq.~\eqref{eq:optimal_base_stock} that uses two scalar quantities, $z^*$ and $\tilde{I}_t$, as inputs. Hence, the bias term $e_1\in\mathbb{R}$ in the first layer is a scalar that functions as an analog of $z^*$ and the corresponding weight vector $\mathbf{w}_1\in \mathbb{R}^{l+1}$ is used to transform the input $\mathbf{s}_t$ into an analog of $\tilde{I}_t$. Note that although we select the minimal architecture that captures the basestock structure, we do not preset $\mathbf{w}_1=-\mathbf{1}$ but rather let all weights be determined via training. This is desirable because in dual-sourcing problems, where the optimal structure is unknown, the network should have flexibility to set the weights to learn more complicated relations. 

After computing $y_1$, this quantity is passed into the CELU activation function that outputs

\begin{equation}
y_2=w_2{\rm CELU}(y_1,\alpha)\,,
\end{equation}
where $w_2\in\mathbb{R}$ denotes a weight that is applied to the CELU output. In single-sourcing problems, the output of the neural network is a single order quantity; hence, $y_2$ is a scalar quantity. To obtain non-negative integer-valued order quantities, we employ fractional decoupling as described in Section~\ref{sec:frac_decoupling}. We train the neural network minimizing $\hat{J}^{(\hat{\pi}_t)} $ [see Eq.~\eqref{eq:emp_loss}] for a discount factor $\gamma=1$ and minibatch size of $M=128$ using the RMSprop optimizer (see the e-companion for further details).
\subsection{Learning an Optimal Base-Stock Policy}
For a concrete example that illustrates how NNCs approximate an optimal base-stock policy \eqref{eq:optimal_base_stock} over a training time horizon of $T=50$, we set $h=5$, $b=495$, $l=0$, and we use a uniform demand distribution $\mathcal{U}\{0,4\}$. In this example, there is only one input $s_t=I_t$ and one weight $w_1$ in the first layer. We represent the optimal inventory position $z^*$ in Eq.~\eqref{eq:optimal_base_stock} by a bias term $e_1$. The outputs of the first and second layers are 
\begin{align}
y_1=w_1 s_t+e_1\quad\mathrm{and}\quad y_2=w_2 {\rm CELU}(y_1,\alpha)\,,
\end{align}
respectively. To obtain a minimum-size representation of \eqref{eq:optimal_base_stock}, we do not include another bias term that is applied to $y_2$. The overall output of the neural network is given by Eq.~\eqref{eq:fract_decoupling}.

For the given parameters, the optimal order-up-to level is $z^*=4$ and the corresponding optimal expected cost per period is $h(z^*-\bar{D})=10$, where $\bar{D}=2$ is the mean demand in one period. We find that the employed NNC approaches the expected cost level of the optimal base-stock policy after about 3,000 training epochs [solid blue line in Figure~\ref{fig:base_stock}(b)]. The total training time is about 1.5 minutes on a regular personal computer. While approaching the optimal solution, small changes of the neural-network weights may entail large changes in the expected cost, leading to the onset of oscillatory behavior of the expected cost after about 3,000 training epochs. If we add a rounding layer instead of using fractional decoupling, the neural network does not reach the optimal base-stock cost due to the reduced set of possible gradient-descent directions [solid red line in Figure~\ref{fig:base_stock}(b)]. 

After training, we selected the best model and extracted the weight and bias values. We find that $w_1=-0.5328$, $e_1=2.1352$, and $w_2=1.8739$. Notice that the NNC approximates the optimal base-stock policy \eqref{eq:optimal_base_stock} since $-e_1/w_1\approx 4= z^*$ and $-w_2 w_1\approx 1$. Figures \ref{fig:base_stock}(c,d) show that the NNC we use in this example learned to produce order, inventory, and cost profiles that resemble those of an optimal base-stock policy. For cases with positive lead times, we obtain similar results, which we omit for the sake of brevity since the optimal solution has the same structure.
\section{Managing Dual-Sourcing Problems with Recurrent Neural Networks}
After having described the basic mechanisms of NNCs in the context of inventory dynamics with a single supplier, we will now focus on dual-sourcing dynamics as a more complex example of inventory-management problems.
\label{sec:nnc_inventory}
\subsection{Dual-Sourcing Model}
\label{sec:model}
We consider a dual-sourcing inventory control problem with stochastic demand. The first sourcing option is a ``regular'' supplier, ${\rm R}$, that delivers goods with an integer lead time $l_{\rm r}>0$ at a cost $c_{\rm r}$. Goods can also be ordered from a second ``emergency'' supplier, ${\rm E}$, with a shorter (non-negative) integer lead time $l_{\rm e}<l_{\rm r}$ at a higher cost $c_{\rm e}>c_{\rm r}$. The premium for the expedited delivery through supplier ${\rm E}$ is thus $c \coloneqq c_{\rm e}-c_{\rm r}>0$.

We use $I_t$ to denote the net inventory at the beginning of period $t\in\{1,2,\dots\}$. The corresponding replenishment orders placed to suppliers ${\rm E}$ and ${\rm R}$ in period $t$ are $q_t^{\rm e}$ and $q_t^{\rm r}$, respectively. In each period $t$, the demand $D_t$ is i.i.d.\ distributed according to a distribution function $\phi$ with finite mean $\mu$.

Using these definitions, the sequence of events in each period of the dual-sourcing model are as follows.
\begin{enumerate}
    \item At the beginning of period $t$, the inventory manager places replenishment orders $q_t^{\rm r}$ and $q_t^{\rm e}$ based on the last-observed net inventory, $I_{t}$, and the replenishment orders that have not arrived yet, $\mathbf{Q}^{\rm r}_{t}=(q^{\rm r}_{t-l_{\rm r}},\dots,q^{\rm r}_{t-1})$ and $\mathbf{Q}^{\rm e}_{t}=(q^{\rm e}_{t-l_{\rm e}},\dots,q^{\rm e}_{t-1})$.
    \item Orders $q_{t-l_{\rm r}}^{\rm r}$ and $q_{t-l_{\rm e}}^{\rm e}$ arrive and are added to the current inventory.
    \item The demand $D_t\sim \phi$ is revealed and subtracted from the current inventory.
\end{enumerate}
Hence, the inventory (with backlogged excess demand) evolves according to
\begin{equation}
I_{t+1}=I_{t}+q^{\rm r}_{t-l_{\rm r}}+q^{\rm e}_{t-l_{\rm e}}-D_t\,. \label{eq:inv_update}
\end{equation}
The dual-sourcing problem is a Markov decision process with state $\mathbf{s}_{t}=(I_{t},\mathbf{Q}_{t}^{\rm r},\mathbf{Q}_{t}^{\rm e})$. For an action $a_{t}=(q_t^{\rm r},q_t^{\rm e})$, the corresponding total cost in period $t$ is
\begin{equation}
c_t(\mathbf{s}_t,\mathbf{a}_{t})=c_{\rm r}q^{\rm r}_t+c_{\rm e}q^{\rm e}_t+h[I_t+q^{\rm r}_{t-l_{\rm r}}+q^{\rm e}_{t-l_{\rm e}}-D_t]^+ + b[D_t-I_t-q^{\rm r}_{t-l_{\rm r}}-q^{\rm e}_{t-l_{\rm e}}]^+\,,
\label{eq:cost_function}
\end{equation}
where $[x]^+=\max(0,x)$, and $h$ and $b$ are the holding and shortage costs, respectively.

It is well-known that Eq.~\eqref{eq:average_exp_cost} admits a stationary optimal policy $\pi^*$ for dual sourcing when the demand distribution has a finite support over the time horizon $T$ \citep{hua2015structural}. The general optimal policy for dual-sourcing problems has not been characterized analytically. In the e-companion, we describe the value iteration that we use to compute optimal order policies for small-scale dual-sourcing instances.
\subsection{Approximating Optimal Dual-Sourcing Policies}
Our goal is to design NNCs that are able to solve dual-sourcing inventory management problems that have no known analytical solution. Similar to the neural-network architecture used in the single-sourcing problem, we employ a neural network that uses as input the ${(l_{\rm r}+l_{\rm e}+1)}$-dimensional state $\mathbf{s}_t=(I_{t},\mathbf{Q}_{t}^{\rm r},\mathbf{Q}_{t}^{\rm e})$ [Figure~\ref{fig:dual_sourcing}(a)]. After processing the inputs in a set of hidden layers, the neural network outputs the action $\hat{\mathbf{a}}_t=(\hat{q}_t^{\rm r},\hat{q}_t^{\rm e})$. Note that actions that are generated by a neural network depend on both the state $\mathbf{s}_t$ and neural network weights $\mathbf{w}$, that is, $\hat{\mathbf{a}}_t=\hat{\mathbf{a}}_t(\mathbf{s}_t;\mathbf{w})$. For training the neural network, we calculate the expected total cost over $T$ periods using Eqs.~\eqref{eq:emp_loss} and \eqref{eq:cost_function}. Neural network weights and biases, indicated by ``+1s'' in Figure~\ref{fig:dual_sourcing}(a), are adjusted by backpropagating gradients that result from minimizing the expected cost. As in the previous section, we backpropagate gradients using fractional decoupling [see Eq.~\eqref{eq:fract_decoupling}]. The main steps of the training algorithm are summarized in Algorithm~\ref{alg:training}. In lines \ref{ln:start_loop}-\ref{ln:end_loop}, the algorithm evaluates an ordering policy implied by the neural network weights $\mathbf{w}^{(n)}$ in epoch $n$. The outer loop in lines \ref{ln:start_outer}-\ref{ln:end_outer} performs the order optimization using gradient descent on the neural network weights $\mathbf{w}^{(n)}$. All neural networks that we use in our numerical experiments consist of seven fully connected, hidden layers with 128, 64, 32, 16, 8, 4, and 2 CELU neurons, respectively. For more details on the network structure and learning dynamics, we refer the reader to the e-companion.

\begin{algorithm}
\caption{A generic algorithm that describes a simplified training procedure of NNCs for controlling dual-sourcing problems. An NNC represents the order policy $\hat{\pi}$ that depends on the neural-network parameters $\mathbf{w}$.}\label{alg:training}
\begin{algorithmic}[1]
\Inputs{$\phi$, $c_{\rm r}$, $c_{\rm e}$, $l_{\rm r}$, $l_{\rm e}$, $h$, $b$, $T$, $M$, $\eta$, $\Call{nn\_controller}$, $\Call{inventory\_dynamics}$
}
\Outputs{best\_loss, $\mathbf{w}^*$}
\Initialize{$\mathbf{w}^{(1)}$, $\mathbf{s}_1^{(1)},\dots,\mathbf{s}_1^{(M)}$}

\For{n = 1 to max\_epochs} \Comment{Iterate over all training epochs.} \label{ln:start_outer}
    \State total\_cost $\gets$ 0
    \For{t = 1 to T} \Comment{Iterate over discrete time steps in a time horizon $[1,T]$.}            \label{ln:start_loop}
        \State mean\_cost $\gets$ 0
        \For{m = 1 to M} \Comment{Iterate over all samples in a mini-batch.}
                \State $D_t^{(m)}\sim\phi$ \Comment{Sample from a demand distribution, \eg, a uniform distribution.}
                \State $\hat{q}_t^{\rm r}$, $\hat{q}_t^{\rm e}\gets  \Call{nn\_controller}{\mathbf{s}_t^{(m)}, \mathbf{w}^{(n)}}$ \Comment{NN controller calculates current orders.}
                \State $\mathbf{s}_{t+1}^{(m)},c_t^{(m)} \gets \Call{inventory\_dynamics}{D_t^{(m)}, \mathbf{s}_t^{(m)}, \hat{q}_t^{\rm r}, \hat{q}_t^{\rm e}}$ \Comment{State update (\ref{eq:inv_update}, \ref{eq:cost_function}).}
                \State mean\_cost += $c_t^{(m)}/M$ \Comment{Calculate mean cost over all samples per time step.}
        \EndFor \label{ln:end_loop}
    \State total\_cost +=  mean\_cost
    \EndFor 

    \State  $J^{(\hat{\pi}_t)} \gets$  total\_cost \Comment{Set the learning loss as the expected total cost over all samples.}

    \State $\mathbf{w}^{(n+1)}\gets\mathbf{w}^{(n)}-\eta \nabla_{\mathbf{w}^{(n)}} 
J^{(\hat{\pi}_t)}$ \Comment{Perform a gradient descent step for NN parameters.}

\Comment{Other parameter update methods (\eg, RMSProp) can be used in this step as well.}

    \If{total\_cost $<$ best\_loss} \Comment{Check for best performance over epochs $1,\dots,n$.}
        \State $\mathbf{w}^* \gets \mathbf{w}^{(n+1)}$ \Comment{Save best neural-network parameters $\mathbf{w}^*$.}
        \State best\_loss $\gets$ total\_cost
    \EndIf
\EndFor \label{ln:end_outer}
\State \Return best\_loss, $\mathbf{w}^*$
\end{algorithmic}
\end{algorithm}
\section{Computational Experiments}
\label{sec:computations}
We carry out computational experiments to shed light on a matter of aspects pertinent to our method's performance, which we organize in four parts. The first part demonstrates performance in terms of the expected cost per period, when compared to the optimal solution and to an alternative, state-of-the-art benchmark. In addition, we illustrate to what extent NNCs approximate optimal solutions not only in the sense of obtaining near-optimal cost values, but also in how these values are distributed in a variety of samples. The second part extends the analysis of the first part to a larger set of instances, and compares the neural network control approach with the optimal solutions and the alternative benchmark. We apply transfer learning to take advantage of existing ordering policies and warm-start the neural network training. Transfer learning can help to reduce the number of training epochs by a factor between 10 and 30, depending on the instance. Moreover, in this part we study the ability of NNCs to approximate optimal solutions in a stronger sense, namely its ability to generate ordering policies that are close to the optimal ones. To show that NNCs are also applicable in scenarios where the optimal policy structure is not known and no good heuristics are available, the third part focuses on dual sourcing with fixed order costs~\citep{svoboda2021typology}. Finally, the fourth part applies neural network control on a subset of real demand data, obtained by \cite{manary2021data}, and shows how it can outperform an alternative approach. In this last part, we also compare the performance of different neural-network structures, including long short-term memory (LSTM)~\citep{DBLP:journals/neco/HochreiterS97} and transformer architectures~\citep{vaswani2017attention}.

In all our experiments on dual-sourcing problems (without fixed order costs), we compare the neural network's performance with the capped dual index (CDI) heuristic that first appeared in \cite{sun2019robust}. The CDI policy is an appropriate benchmark for several reasons. First, it can be seen as a generalized version of the tailored base-surge heuristic, which is asymptotically optimal for large $l_{\rm r}$ \citep{xin2018asymptotic}. Second, it is easy to understand, implement and communicate, and as such it is an appropriate baseline for a more complex approach. Third, it has state-of-the-art performance, which, combined with its simplicity (only 3 parameters need to be optimized), make it favorable against more complex approaches that may outperform it marginally. Finally, \cite{gijsbrechts2020can} showed that CDI is able to outperform state-of-the-art RL policies.

Experiments were carried out on an AMD\textsuperscript{\textregistered} Ryzen threadripper 3970. In this machine, 100 training epochs for $T=100$ periods and a minibatch size $M=512$ take about 20 seconds to complete. Without having access to a pretrained neural network, obtaining optimality gaps of about 0.1\% requires between 2,000 and 3,000 training epochs, which corresponds to less than 10 minutes of CPU time. Pretrained neural networks can achieve good performance on unseen instances after training them for a few hundred epochs. For comparison, dynamic programming solutions of certain instances took about 50 days of computing time. All implementations are available on Gitlab~\citep{GitLab}. Further details on implementation attributes of the underlying algorithms are also summarized in the e-companion. Additionally, we show in the e-companion that the runtime increases almost linearly from $T=100$ to $T=1100$.

\subsection{Visualizing NNC performance}
\label{sec:NNC_performance}
In this part, we use nine small instances with discrete uniform demand distribution $\mathcal{U}\{0,4\}$, and parameters $h=5$, $b=495$, $c_{\rm e} \in \{5, 10, 20\}$, and $l_{\rm r}\in \{2, 3, 4\}$. These parameters have been chosen in accordance with \cite{gijsbrechts2020can} and \cite{scheller2007effective}. In all experiments, we set $c_{\rm r}=0$ and $l_{\rm e}=0$ without loss of generality \citep{sheopuri2010new}. The latter assumption is justified because any system with $l_{\rm e}>0$ can be transformed to an equivalent one with $l_{\rm e}=0$ \citep{sun2019robust, xin20211}. Such small instances have been used in the literature to evaluate various heuristics \citep{scheller2007effective, gijsbrechts2020can}. Their advantage is that exact optimal solutions can be obtained, and therefore one can evaluate performance in a formal sense. We use these instances to investigate the performance and structure of solutions obtained by our approach. A potential limitation of such instances is that it is not clear whether a method that performs well in such small instances will perform well in more realistic demand distributions. We address this point using real demand data in the third part of our analysis.

\begin{figure}
    \FIGURE
    {\includegraphics[width=\textwidth]{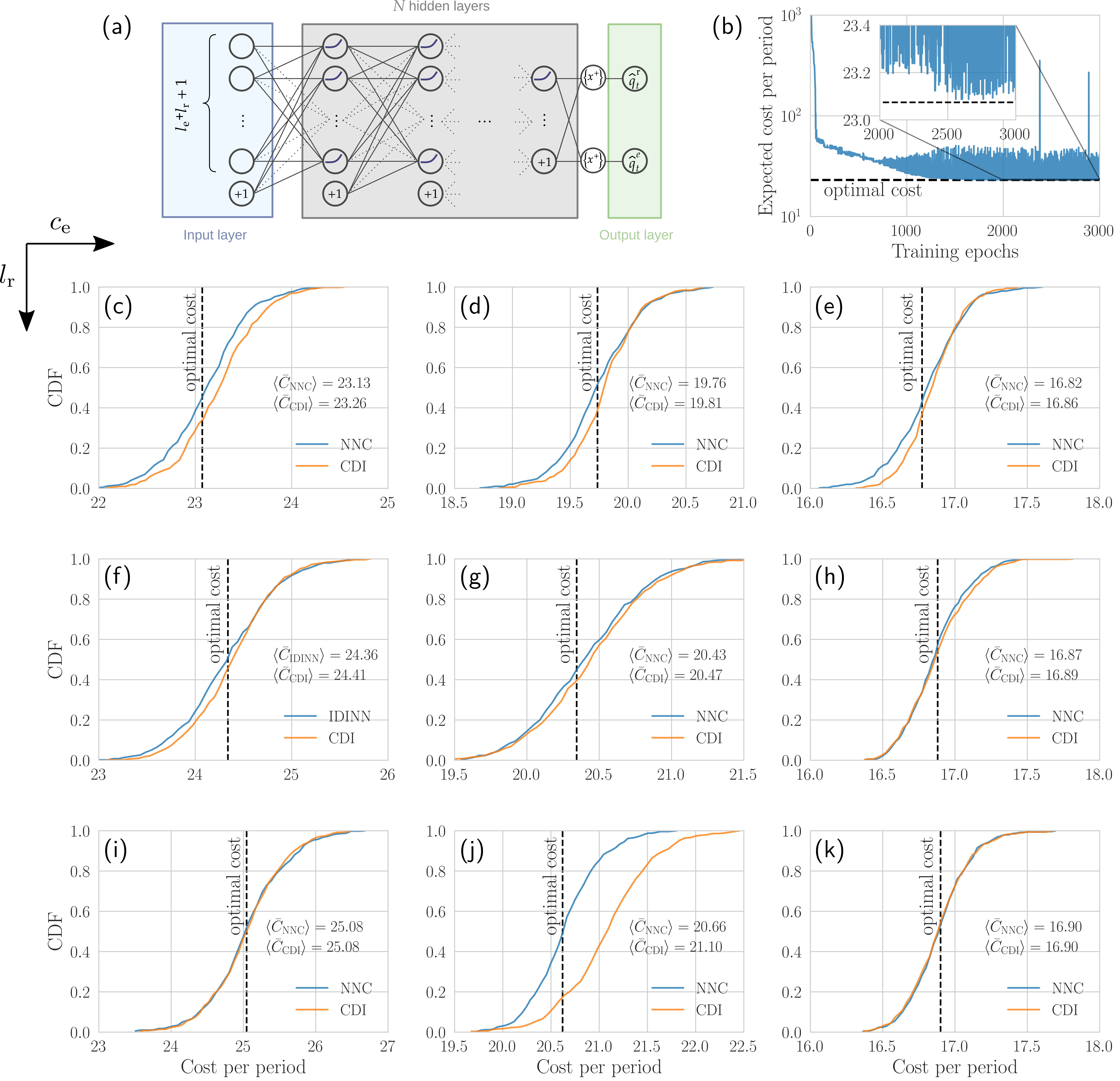}}
    {Training of NNC and expected cost distribution for dual-sourcing inventory systems.\label{fig:dual_sourcing}}
    {(a) Schematic of an NNC architecture for dual sourcing. The neural network uses as input the ${(l_{\rm r}+l_{\rm e}+1)}$-dimensional state $\mathbf{s}_t=(I_{t},\mathbf{Q}_{t}^{\rm r},\mathbf{Q}_{t}^{\rm e})$ and outputs an action $\hat{a}_t=(\hat{q}_t^{\rm r},\hat{q}_t^{\rm e})$. Hidden layers are composed of CELU activations and bias terms, indicated by ``+1''s. (b) Expected cost per period as a function of training epochs for $h=5$, $b=495$, $c_{\rm r}=0$, $c_{\rm e}=20$, $l_{\rm r}=2$, $l_{\rm e}=0$, $T=100$, and demand distribution $\mathcal{U}\{0,4\}$. The dashed black line indicates the approximately optimal cost value found with a dynamic programming approach. (c--k) Cumulative distribution functions (CDFs) of the cost per period for CDI and NNC policies. The shown distributions are based on $500$ realizations, each consisting of a time horizon $T=1,000$. Expedited order costs are 20, 10, and 5 (from left to right); regular order lead time are 2, 3, and 4 (from top to bottom). The expected costs per period associated with NNC and optimized CDI policies are denoted $\langle \bar{C}_{\rm NNC} \rangle$ and $\langle \bar{C}_{\rm CDI} \rangle$, respectively. Order policies that are based on NNCs stochastically dominate CDI in seven out of nine instances. In the remaining two cases, the performance is almost identical.}
\end{figure}

Panel (a) of Figure~\ref{fig:dual_sourcing} shows the generic design of the network, while panel (b) shows the evolution of the expected cost per period during training for $l_{\rm r}=2$, $c_{\rm e}=20$, and $T=100$. We observe that the cost reaches values below 100 after a few training epochs and approaches the optimal solution after about 2,500 epochs (\ie, after a few minutes of CPU time).
Panels (c--k) show the cumulative distribution functions (CDFs) of the expected costs per period for $c_{\rm e}=20,10,5$ (from left to right) and $l_{\rm r}=2,3,4$ (from top to bottom). Distributions are formed by $500$ realizations, each consisting of a time horizon $T=1,000$. The dashed black line indicates the corresponding infinite horizon optimal expected cost per period (numerical values are reported in Table~\ref{tab:dual_sourcing_costs}). We observe that NNC order policies stochastically dominate optimized CDI policies in seven of out nine instances. For the remaining two instances, NNC policies have an expected cost that matches that of CDI. In conclusion, these experiments imply that NNC has comparable\textemdash and even better\textemdash performance to CDI. The absolute cost difference is small, because both methods are near optimal. However, what is perhaps unexpected is the consistency by which NNC outperforms CDI, not only in terms of expected cost, but also on a realization-by-realization basis, as illustrated in panels (c--k) of Figure~\ref{fig:dual_sourcing}.
\subsection{Learning Optimal Solutions}
\label{sec:oc_solutions}
We now extend the set of instances to include two backlog cost levels ($b\in \{95, 495
\}$) and two demand distributions ($\mathcal{U}\{0,4\}$ and $\mathcal{U}\{0,8\}$). The NNC, CDI, and DP expected costs per period of these instances are summarized in Table~\ref{tab:dual_sourcing_costs}. The mean expected costs per period of NNCs and CDI are based on $500$ realizations, each consisting of a time horizon $T=1,000$. During training, we use shorter time horizons between $100$ and $200$ periods (see the e-companion for further details). In all 36 instances, the cost of NNC order policies is lower than or equal to that of corresponding optimized CDI policies. For the majority of neural networks that we used to generate the results shown in Tab.~\ref{tab:dual_sourcing_costs}, we employed a transfer-learning approach~\citep{bozinovskiinfluence} to speed-up training. We found that with transfer learning some neural networks reached a good performance after a few hundred training epochs instead of after a few thousand training epochs, which corresponds to training times of less than 1 minute. The holding and backlog costs that we use in our numerical experiments correspond to service levels $b/(h+b)$ of 95 and 99\%. Because back orders occur more frequently in optimal order policies for lower service levels, one can expect that it is more difficult for NNCs to learn optimal replenishment orders. In the e-companion, we show that NNCs also outperform CDI policies or achieve equal performance for a lower service level of 85\% that has been used by \cite{sun2019robust} to study the performance of CDI. 

In order to quantify the similarity between (i) the optimal policy and (ii) policies that are based on NNC and CDI, we measure the corresponding root-mean-square error (RMSE). Specifically, if $\mathcal{S}_{\rm DP}$ is the set of recurrent states in the obtained optimal solution, we define the RMSE of method $m$ as 
\begin{align}
    {\rm{RMSE}}_m = \sqrt{\frac{1}{\lvert \mathcal{S}_{\rm{DP}} \rvert} \sum_{s \in\mathcal{S}_{\rm{DP}}} \left(q_{\rm {DP}}^{\rm{r}}(\mathbf{s})-q_m^{\rm{r}}(\mathbf{s})\right)^2 + \left(q_{\rm{DP}}^{\rm{e}}(\mathbf{s})-q_m^{\rm{e}}(\mathbf{s})\right)^2}\,,
\end{align}
where $q_{\rm {DP}}^{\rm{r}}(\mathbf{s})$ and $q_{\rm {DP}}^{\rm{e}}(\mathbf{s})$ respectively denote the optimal (``dynamic program'') regular and expedited orders associated with state $\mathbf{s}$. The corresponding regular and expedited orders of method $m$ are $q_m^{\rm{r}}(\mathbf{s})$ and $q_m^{\rm{e}}(\mathbf{s})$, respectively. In our experiments, $\mathcal{S}_{\rm DP}$ is a subset of the recurrent states of both CDI and NNC. To this end, the last two columns of Table~\ref{tab:dual_sourcing_costs} show the corresponding RMSEs. We see that NNCs have a lower average and median RMSE compared to CDI (0.51 vs 0.59; 0.50 vs 0.57, respectively); NNCs attain a lower RMSE in more than half the instances. It should be noted that NNCs are trained using the ${(l_{\rm r}+1)}$-dimensional state $\mathbf{s}_t=(I_{t},\mathbf{Q}_{t}^{\rm r})$, while the states of the DP and CDI are in the compact, $l_{\rm r}$-dimensional space which considers the net inventory \textit{after} replenishment, namely $\mathbf{s}_t^{\rm compact}=(I_t+q_{t-l_{\rm r}}^{\rm r}, q_{t-l_{\rm r}+1}^{\rm r}, \dots, q_{t-1}^{\rm r})$. When we convert the NNC ordering policies in the compact space, we average the corresponding ordering quantities. The resulting fractional ordering quantities can be interpreted as randomized ordering policies.

\begin{figure}
    \FIGURE
    {\includegraphics{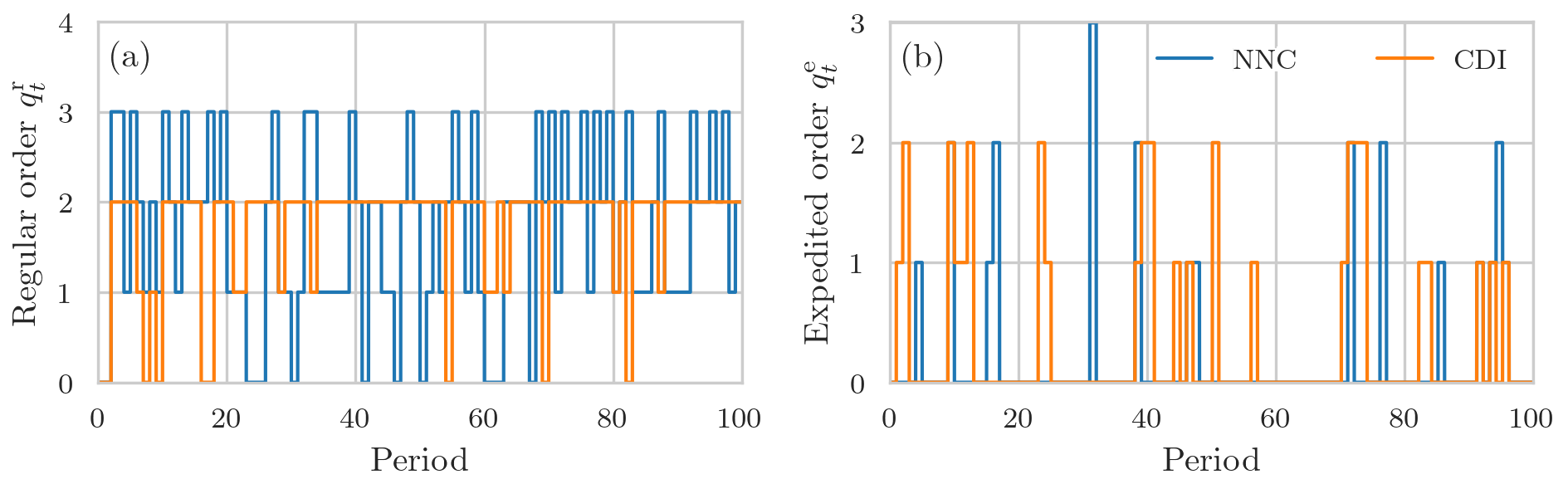}}
    {Comparison of NNC and CDI orders.\label{fig:NNC_cdi_orders}}
    {Evolution of NNC and CDI regular orders (a) and expedited orders (b). The parameters of the underlying dual-sourcing problem are $h=5$, $b=495$, $c_{\rm r}=0$, $c_{\rm e}=20$, $l_{\rm r}=2$, and $l_{\rm e}=0$. The demand distribution is $\mathcal{U}\{0,4\}$.}
\end{figure}

Figure~\ref{fig:NNC_cdi_orders} shows an example of the evolution of regular and expedited orders that are based on NNC and CDI policies. One advantage of NNC policies over CDI is that regular orders can be placed in a way that is tailored to the current inventory and previously placed orders. As a result, fewer expedited orders need to be placed to replenish inventory.

\begin{table}[hbt!]
\TABLE
{Expected costs per period of CDI and NNC order policies for dual sourcing.\label{tab:dual_sourcing_costs}}
{\centering
\renewcommand*{\arraystretch}{1.2}
\begin{tabular}{ >{\centering\arraybackslash} m{4em} >{\centering\arraybackslash} m{4em}
>{\centering\arraybackslash} m{4em} 
>{\centering\arraybackslash} m{4em} 
>{\centering\arraybackslash} m{4em} 
>{\centering\arraybackslash} m{4em} 
>{\centering\arraybackslash} m{4em}
>{\centering\arraybackslash} m{4em}
>{\centering\arraybackslash} m{4em}}\toprule
$l_{\rm r}$ & $c_{\rm e}$ & $b$ & demand & CDI cost & NNC cost & DP cost & CDI RMSE & NNC RMSE
\\[1pt] \hline\hline
2 & 5 & 95 & $\mathcal{U}\{0,4\}$ & 16.87 & \textbf{16.80} & 16.77 & 0.58 & \textbf{0.56}\\[1pt]
2 & 5 & 95 & $\mathcal{U}\{0,8\}$ & 32.41 & \textbf{32.33} & 32.27& \textbf{0.31} & 0.57\\[1pt]
2 & 5 & 495 & $\mathcal{U}\{0,4\}$ & 16.86 & \textbf{16.82} & 16.77 & 0.58 & \textbf{0.48}\\[1pt]
2 & 5 & 495 & $\mathcal{U}\{0,8\}$ & \textbf{32.28} & \textbf{32.28} & 32.27& \textbf{0.31} & 0.50\\[1pt]
2 & 10 & 95 & $\mathcal{U}\{0,4\}$ & 19.81 & \textbf{19.79} & 19.73 & \textbf{0.40} & 0.56\\[1pt]
2 & 10 & 95 & $\mathcal{U}\{0,8\}$ & 37.42 & \textbf{37.24} & 37.24& 0.57 & \textbf{0.45}\\[1pt]
2 & 10 & 495 & $\mathcal{U}\{0,4\}$ & 19.81 & \textbf{19.76} & 19.74 & 0.40 & \textbf{0.26}\\[1pt]
2 & 10 & 495 & $\mathcal{U}\{0,8\}$ & \textbf{37.92} & \textbf{37.92} & 37.84& 0.60 & \textbf{0.52}\\[1pt]
2 & 20 & 95 & $\mathcal{U}\{0,4\}$ & 23.01 & \textbf{22.99} & 22.83& 0.54 & \textbf{0.50}\\[1pt]
2 & 20 & 95 & $\mathcal{U}\{0,8\}$ & 41.73 & \textbf{41.68} & 41.64& 0.65 & \textbf{0.46}\\[1pt]
2 & 20 & 495 & $\mathcal{U}\{0,4\}$ & 23.26 & \textbf{23.13} & 23.07 & 0.46 & \textbf{0.45}\\[1pt]
2 & 20 & 495 & $\mathcal{U}\{0,8\}$ & 43.82 & \textbf{43.79} & 43.77& 0.54 & \textbf{0.40}\\[1pt]\midrule
3 & 5 & 95 & $\mathcal{U}\{0,4\}$ & \textbf{16.88} & \textbf{16.88} & 16.88 & 0.53 & \textbf{0.43}\\[1pt]
3 & 5 & 95 & $\mathcal{U}\{0,8\}$ & 32.93 & \textbf{32.65} & 32.60& 0.86 & \textbf{0.49}\\[1pt]
3 & 5 & 495 & $\mathcal{U}\{0,4\}$ & 16.89 & \textbf{16.87} & 16.88 & 0.53 & \textbf{0.44}\\[1pt]
3 & 5 & 495 & $\mathcal{U}\{0,8\}$ & 32.80 & \textbf{32.66} & 32.60& \textbf{0.86} & 0.93\\[1pt]
3 & 10 & 95 & $\mathcal{U}\{0,4\}$ & 20.48 & \textbf{20.40} & 20.34& \textbf{0.36} & 0.37\\[1pt]
3 & 10 & 95 & $\mathcal{U}\{0,8\}$ & 38.79 & \textbf{38.69} & 38.64& 0.64 & \textbf{0.58}\\[1pt]
3 & 10 & 495 & $\mathcal{U}\{0,4\}$ & 20.47 & \textbf{20.43} & 20.34 & \textbf{0.36} & 0.77\\[1pt]
3 & 10 & 495 & $\mathcal{U}\{0,8\}$ & 39.10 & \textbf{38.97} & 38.89& 0.97 & \textbf{0.77}\\[1pt]
3 & 20 & 95 & $\mathcal{U}\{0,4\}$ & 24.44 & \textbf{24.43} & 24.30 & 0.57 & \textbf{0.49}\\[1pt]
3 & 20 & 95 & $\mathcal{U}\{0,8\}$ & 44.70 & \textbf{44.59} & 44.44& \textbf{0.61} & \textbf{0.61}\\[1pt]
3 & 20 & 495 & $\mathcal{U}\{0,4\}$ & 24.41 & \textbf{24.36} & 24.34 & 0.44 & \textbf{0.33}\\[1pt]
3 & 20 & 495 & $\mathcal{U}\{0,8\}$ & 46.39 & \textbf{46.33} & 46.20& \textbf{0.57} & 0.63\\[1pt]\midrule
4 & 5 & 95 & $\mathcal{U}\{0,4\}$ & \textbf{16.90} & \textbf{16.90} & 16.90 & 0.57 & \textbf{0.51} \\[1pt]
4 & 5 & 95 & $\mathcal{U}\{0,8\}$ & 32.95 & \textbf{32.82} & 32.71 & 0.80 & \textbf{0.62} \\[1pt]
4 & 5 & 495 & $\mathcal{U}\{0,4\}$ & \textbf{16.90} & \textbf{16.90} & 16.90 & 0.57 & \textbf{0.50}\\[1pt]
4 & 5 & 495 & $\mathcal{U}\{0,8\}$ & 32.94 & \textbf{32.72} & 32.72 & 0.80 & \textbf{0.58} \\[1pt]
4 & 10 & 95 & $\mathcal{U}\{0,4\}$ & 21.10 & \textbf{20.69} & 20.61 & 0.55 & \textbf{0.44}\\[1pt]
4 & 10 & 95 & $\mathcal{U}\{0,8\}$ & 39.63 & \textbf{39.49} & 39.25 & 1.10 & \textbf{0.60} \\[1pt]
4 & 10 & 495 & $\mathcal{U}\{0,4\}$ & 21.10 & \textbf{20.66} & 20.61 & 0.55 & \textbf{0.38}\\[1pt]
4 & 10 & 495 & $\mathcal{U}\{0,8\}$ & 39.63 & \textbf{39.45} & 39.35 & 0.86 & \textbf{0.45} \\[1pt]
4 & 20 & 95 & $\mathcal{U}\{0,4\}$ & \textbf{25.08} & \textbf{25.08} & 24.56 & \textbf{0.36} & 0.51\\[1pt]
4 & 20 & 95 & $\mathcal{U}\{0,8\}$ & 46.46 & \textbf{46.14} & 46.02 & 0.84 & \textbf{0.44}\\[1pt]
4 & 20 & 495 & $\mathcal{U}\{0,4\}$ & \textbf{25.08} & \textbf{25.08} & 25.04 & \textbf{0.32} & 0.37\\[1pt]
4 & 20 & 495 & $\mathcal{U}\{0,8\}$ & 47.66 & \textbf{47.56} & 47.53 & 0.63 & \textbf{0.39} \\[1pt]\bottomrule

\end{tabular}
\vspace{1mm}}
{CDI and NNC costs are based on $500$ realizations, each consisting of a time horizon $T=1,000$. For comparison, we also show the corresponding optimal, infinite-horizon DP cost. The parameters $l_{\rm r}$, $c_{\rm e}$, and $b$ and the demand distribution are as listed in the first four columns. The remaining parameters are held constant at $c_{\rm r} = 0$ and $h = 5$.}
\end{table}
\subsection{Dual Sourcing with Fixed Costs}
\label{sec:fixed_costs}
In this section, we investigate the effectiveness of NNC policies on dual-sourcing problems that have fixed costs. Specifically, we assume that both the expedited and the regular supplier charge a fixed amount per order, denoted $f_{\rm r}$ and $f_{\rm e}$, respectively. As for the order costs $c_{\rm r}$ and $c_{\rm e}$, one typically imposes $f_{\rm r} < f_{\rm e}$. Such problems have appeared sparingly in the inventory management literature~\citep{svoboda2021typology} because key properties of single-sourcing models, such as quasi-concavity and K-convexity, do not hold for problems with multiple sources~\citep{fox2006optimal}. Therefore, much of the research has focused on special cases. For example, \cite{fox2006optimal} identified the optimal policies when one supplier has negligible variable costs, the other supplier negligible fixed costs, and the demand distribution is log-concave; \cite{axsater2007heuristic} proposes a heuristic for triggering emergency orders under compound Poisson demand; ~\cite{huggins2010inventory} focus on the case where expedited orders occur only when there are shortages; and \cite{johansen2014emergency} assume a fixed order quantity from the regular supplier. Table~\ref{tab:dual_sourcing_fixed_costs} shows both the expected costs per period of NNC policies and the corresponding optimal values obtained with a dynamic-programming approach.
\begin{table}[hbt!]
\TABLE
{Expected costs per period of NNC order policies for dual sourcing with fixed order cost.\label{tab:dual_sourcing_fixed_costs}}
{\centering
\renewcommand*{\arraystretch}{1.2}
\begin{tabular}{ >{\centering\arraybackslash} m{4em} >{\centering\arraybackslash} m{4em}
>{\centering\arraybackslash} m{4em} 
>{\centering\arraybackslash} m{4em} 
>{\centering\arraybackslash} m{4em} 
>{\centering\arraybackslash} m{4em}}\toprule
$l_{\rm r}$ & $c_{\rm e}$ & $b$ & demand & NNC cost & DP cost \\[1pt] \hline\hline
2 & 5 & 95 & $\mathcal{U}\{0,4\}$ & 24.36 & 23.61 \\[1pt]
2 & 5 & 95 & $\mathcal{U}\{0,8\}$ & 41.41 & 40.30 \\[1pt]
2 & 5 & 495 & $\mathcal{U}\{0,4\}$ & 24.36 & 23.61 \\[1pt]
2 & 5 & 495 & $\mathcal{U}\{0,8\}$ & 41.71 & 40.96 \\[1pt]
2 & 10 & 95 & $\mathcal{U}\{0,4\}$ & 26.13 & 25.63 \\[1pt]
2 & 10 & 95 & $\mathcal{U}\{0,8\}$ & 44.32 & 43.74 \\[1pt]
2 & 10 & 495 & $\mathcal{U}\{0,4\}$ & 26.61 & 25.90 \\[1pt]
2 & 10 & 495 & $\mathcal{U}\{0,8\}$ & 45.86 & 45.14 \\[1pt]
2 & 20 & 95 & $\mathcal{U}\{0,4\}$ & 27.57 & 26.95 \\[1pt]
2 & 20 & 95 & $\mathcal{U}\{0,8\}$ & 47.45 & 46.76 \\[1pt]
2 & 20 & 495 & $\mathcal{U}\{0,4\}$ & 28.90 & 28.08 \\[1pt]
2 & 20 & 495 & $\mathcal{U}\{0,8\}$ & 50.45 & 49.82 \\[1pt]\midrule
3 & 5 & 95 & $\mathcal{U}\{0,4\}$ & 25.77 & 24.13 \\[1pt]
3 & 5 & 95 & $\mathcal{U}\{0,8\}$ & 42.34 & 41.16\\[1pt]
3 & 5 & 495 & $\mathcal{U}\{0,4\}$ & 25.81 & 24.13 \\[1pt]
3 & 5 & 495 & $\mathcal{U}\{0,8\}$ & 42.39 & 41.73\\[1pt]
3 & 10 & 95 & $\mathcal{U}\{0,4\}$ & 27.89 & 26.75 \\[1pt]
3 & 10 & 95 & $\mathcal{U}\{0,8\}$ & 46.21 & 45.79 \\[1pt]
3 & 10 & 495 & $\mathcal{U}\{0,4\}$ & 27.89 & 26.88 \\[1pt]
3 & 10 & 495 & $\mathcal{U}\{0,8\}$ & 47.29 & 46.86 \\[1pt]
3 & 20 & 95 & $\mathcal{U}\{0,4\}$ & 30.42 & 29.30 \\[1pt]
3 & 20 & 95 & $\mathcal{U}\{0,8\}$ & 51.24 & 50.35 \\[1pt]
3 & 20 & 495 & $\mathcal{U}\{0,4\}$ & 31.17 & 30.07 \\[1pt]
3 & 20 & 495 & $\mathcal{U}\{0,8\}$ & 53.26 & 52.86 \\[1pt]\midrule
4 & 5 & 95 & $\mathcal{U}\{0,4\}$ & 25.90 & 24.31 \\[1pt]
4 & 5 & 95 & $\mathcal{U}\{0,8\}$ & 42.81 & 41.35 \\[1pt]
4 & 5 & 495 & $\mathcal{U}\{0,4\}$ & 25.90 & 24.31 \\[1pt]
4 & 5 & 495 & $\mathcal{U}\{0,8\}$ & 42.81 & 41.89\\[1pt]
4 & 10 & 95 & $\mathcal{U}\{0,4\}$ & 28.42 & 27.27 \\[1pt]
4 & 10 & 95 & $\mathcal{U}\{0,8\}$ & 47.23 & 46.83 \\[1pt]
4 & 10 & 495 & $\mathcal{U}\{0,4\}$ & 28.42 & 27.34 \\[1pt]
4 & 10 & 495 & $\mathcal{U}\{0,8\}$ & 48.00 & 47.74 \\[1pt]
4 & 20 & 95 & $\mathcal{U}\{0,4\}$ & 31.85 & 30.64 \\[1pt]
4 & 20 & 95 & $\mathcal{U}\{0,8\}$ & 53.42 & 52.46 \\[1pt]
4 & 20 & 495 & $\mathcal{U}\{0,4\}$ & 31.90 & 31.15\\[1pt]
4 & 20 & 495 & $\mathcal{U}\{0,8\}$ & 54.96 & 54.59 \\[1pt]\bottomrule

\end{tabular}
\vspace{1mm}}
{NNC costs are based on a time horizon of $T=10,000$. For comparison, we also show the corresponding optimal, infinite-horizon DP cost. The parameters $l_{\rm r}$, $c_{\rm e}$, and $b$ and the demand distribution are as listed in the first four columns. The remaining parameters are held constant at $c_{\rm r} = 0$, $h = 5$, $f_{\rm e}=10$, and $f_{\rm r}=5$. For instances with $l_{\rm  r}=4$ and demand distribution $\mathcal{U}\{0,8\}$, we set the maximum value iteration runtime to 7 days.}
\end{table}
We find that NNC policies produce expected costs that are near-optimal. This is also the case for problems that have low service level (see Table~\ref{tab:dual_sourcing_costs_low_service_fixed_cost} in the e-companion).
\clearpage
\pagebreak

\subsection{Application to Empirical Demand Data}
\label{sec:emp_demand}
\begin{figure}
    \FIGURE
    {\includegraphics{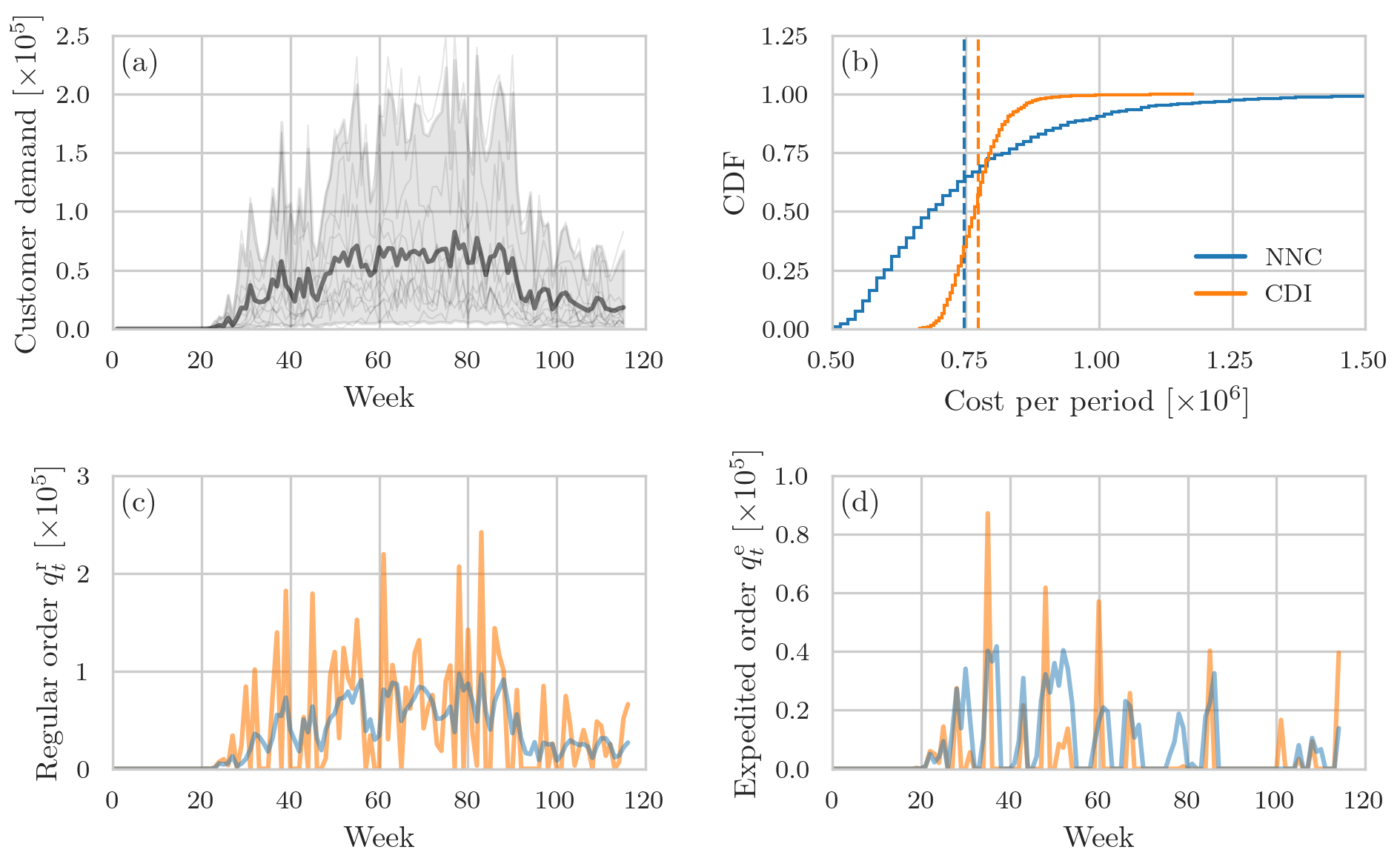}}
    {Controlling real-world inventory management problems using NNCs and CDI.\label{fig:empirical}}
    {(a) Customer demand data (solid grey lines) that is taken from \cite{manary2021data}. The solid black line and grey-shaded regions indicate the mean customer demand and 95\% confidence intervals of the demand distribution \eqref{eq:empirical_demand_distribution}, respectively. (b) The cumulative distribution function of the average NNC and CDI costs. Shown results are based on $10^3$ i.d.d.\ samples. Dashed lines indicate mean expected costs per period [747,117 (NNC) and 773,993 (CDI)]. (c) An example of regular orders generated by an NNC (solid blue line) and a CDI policy (solid orange line). (d) An example of expedited orders generated by an NNC (solid blue line) and a CDI policy (solid orange line). The parameters of the underlying dual-sourcing problem are $h=5$, $b=495$, $c_{\rm r}=0$, $c_{\rm e}=20$, $l_{\rm r}=2$, and $l_{\rm e}=0$.}
\end{figure}

We now apply NNCs to dual-sourcing problems with actual customer demand data taken from \cite{manary2021data}. The data represent customer demand of microprocessors from Intel Corporation. Contrary to the previous datasets, that assumed stationary demand, this dataset models a situation where demand evolves to a peak, then plateaus, and eventually vanishes. Such a demand profile corresponds to the introduction of a new microprocessor generation: at the initial stage, it replaces older microprocessors, then it enjoys a maturity period, and eventually it is replaced by a new generation. Figure~\ref{fig:empirical}(a) shows the evolution of customer demand over $T=115$ weeks. Each grey line represents a different microprocessor generation and indicates a possible evolution of demand. We describe such empirical demand data by a Gaussian process~\citep{roberts2013gaussian} with random demand
\begin{equation}
D_t\sim \mathcal{N}^+(\mu_t,\sigma_t,a,b)\,,\quad t\in\{1,\dots,T\}\,,
\label{eq:empirical_demand_distribution}
\end{equation}
where $\mathcal{N}^+(\mu_t,\sigma_t,a,b)$ denotes the truncated normal distribution with domain $(a,b)$. The quantities $\mu_t$ and $\sigma_t$ denote mean and standard deviation of the empirical demand data at time $t$. The solid black line in Figure~\ref{fig:empirical}(a) shows the evolution of the mean $\mu_t$ and grey-shaded regions indicate the 95\% confidence intervals (CIs). Demands are zero in the beginning of the shown product lifecycle. The peak demand reaches values of about $3\times 10^5$ units after about 70 weeks. To reduce the effect of truncation errors, we chose a relatively broad truncation interval $(a,b)=(0,10^8)$. The key difference between this environment and the archetypal dual-sourcing problem is the finite time horizon. Note that CDI remains robustly optimal for finite time horizons with non-stationary, correlated demand confined to polyhedral uncertainty sets. Therefore, it remains a competitive benchmark, to which we compare the performance of NNCs.

To calculate CDI orders, we determine the order-up-to levels of the regular and expedited suppliers, ${S_t^{\rm r *}}$ and ${S_t^{\rm e *}}$, and the cap ${\bar{q}_t^{\rm r *}}$ (see the e-companion and proposition 3 in \cite{sun2019robust} for further details) by estimating both the minimum and maximum demand using estimates of the 99\% CIs of the distribution $\mathcal{N}^+(\mu_t,\sigma_t,a,b)$. That is,
\begin{align}
\begin{split}
{S_t^{\rm e *}}&={\bar{q}_t^{\rm r *}}=\frac{h [\mu_t-2.58 \sigma_t]^++b [\mu_t+2.58 \sigma_t]}{h+b}\,, \\
{S_t^{\rm r *}}&=\frac{h l [\mu_t-2.58 \sigma_t]^++b l [\mu_t+2.58 \sigma_t]}{h+b}\,,
\end{split}
\label{eq:cdi_estimates}
\end{align}
where $l=l_{\rm r}-l_{\rm e}$. The estimates in Eq.~\eqref{eq:cdi_estimates} are based on the assumption that known moments $\mu_t,\sigma_t$ at time $t$ are used as estimates for times $t+1,\dots,t+l$. In the e-companion, we also study another CDI baseline where we assume that the moments $\mu_t$ and $\sigma_t$ are known exactly up to time $t+l$ so that better estimates of ${S_t^{\rm r *}},{S_t^{\rm e *}},{\bar{q}_t^{\rm r *}}$ can be obtained. In both scenarios, we find that NNC can achieve significantly lower mean and median costs than CDI. In addition to the RNN-based controller, we perform simulations for two additional neural network architectures (LSTMs and transformers) in the e-companion. Although LSTMs and transformers can also achieve relatively small expected costs per period, our results suggest that the inductive bias in the RNN controller (\ie, the use of inventory dynamics to define recurrent connections) helps it to learn more effective order policies.

For a meaningful comparison between NNCs and CDI with parameters as defined in Eq.~\eqref{eq:cdi_estimates}, we let the neural network access $\mu_t$ and $\sigma_t$. Specifically, the neural network input consists of (i) $I_t$, the net inventory; (ii) $\mathbf{Q}_{t}^{\rm r},\mathbf{Q}_{t}^{\rm e}$, the order pipelines associated with slow and fast orders, respectively; and (iii) $\mu_t,\sigma_t$, the mean and standard deviation of the time-varying demand. More details on the employed neural-network structure and training algorithm (``one-shot learning'') are reported in the e-companion.

To compare the performance of NNC and CDI policies for real-world inventory management problems, we generate 1,000 i.i.d.\ samples for $h=5$, $b=495$, $c_{\rm r}=0$, $c_{\rm e}=20$, $l_{\rm r}=2$, and $l_{\rm e}=0$. These Gaussian process samples are different from the ones that we used to train the model. We find that the employed NNC outperforms CDI in more than 65\% of the studied realizations. The median expected costs per period are 692,118 (NNC) and 770,362 (CDI); the mean expected costs per period are 747,117 (NNC) and 773,993 (CDI) [dashed lines in Figure~\ref{fig:empirical}(b)]. We also evaluate the differences between CDI and NNC costs using the Wilcoxon signed-rank test~\citep{wilcoxon1992individual}. Our analysis shows that the null hypothesis that the difference between the CDI and NNC median costs is negative can be rejected ($p<10^{-10}$) in favor of the alternative that the median cost difference is greater than zero. For a further comparison between NNCs and CDI, we set the backlog cost $b=95$ and again analyze the expected costs per period using 1,000 i.i.d.\ samples. We find that the advantage of NNCs over CDI is even more pronounced in this example with a smaller backlog cost. The median expected costs per period are 563,711 (NNC) and 735,265 (CDI) and the mean expected costs per period are 583,873 (NNC) and 736,018 (CDI). In relative terms, the reduction in mean cost is 20\%, which constitutes a significant improvement over the CDI policy.

Figure~\ref{fig:empirical}(c,d) shows an example of regular and expedited orders that are generated by an NNC policy (solid blue lines) and a CDI policy (solid orange lines). We observe that the NNC regular orders fluctuate to a lesser degree between very high and very low orders than those of the employed CDI order policy. Another difference between the shown NNC and CDI policies is that CDI tends to replenish inventory with larger expedited orders. 

To summarize, we have shown that NNCs are able to learn effective order policies that solve inventory management problems with empirical demand data. In the above examples, NNC order policies can lead to savings between 3--20\% compared to an effective state-of-the-art heuristic.
\section{Conclusion}
\label{sec:conclusion}
Our work develops a neural network\textendash based inventory management system that directly learns effective order policies by minimizing the backlogging, holding, and sourcing costs of underlying inventory dynamics.

We have shown that one activation function suffices for neural network controllers (NNCs) to learn the optimal order policy of single-sourcing problems. Extending the structure of the single-sourcing neural network, we developed NNCs that are able to outperform state-of-the-art order heuristics in dual-sourcing problems. We also studied the ability of NNCs to control dual-sourcing problems with fixed costs for which the optimal policy structure is not known and no good heuristics are available. Our numerical results indicate that NNCs are able to achieve near-optimal cost. Finally, we demonstrated the ability of NNCs to control inventory management problems with empirical demand data. In all studied instances, NNCs either outperformed state-of-the-art baselines or achieved equal performance. Training times varied between a few minutes to about one hour on a regular personal computer. All trained neural networks are publicly available (see e-companion) and may be used a starting point for solving related inventory management problems via transfer learning.

Future research may explore the application of neural network\textendash based control methods to partially observable inventory management problems, multi-supplier and multi-echelon problems, different demand distributions, and complex, multi-constraint optimization problems such as effective vaccine distribution~\citep{mak2021managing}. Other promising venues for research include the application of NNCs to continuous-time models~\citep{xin2021dual} and the integration of neural networks that forecast demand~\citep{liu2022iterative} into NNC inventory management systems.
\bibliographystyle{informs2014}
\bibliography{refs.bib}
%
\ECSwitch


%
\ECHead{Electronic companion}
\begin{table}[htb]
\TABLE
{Overview of variables used in the dual-sourcing model and NNCs.\label{tab:model_variables}}
{\centering
\renewcommand*{\arraystretch}{1.6}
\small
\begin{tabular}{ >{\centering\arraybackslash} m{13em}
>{\raggedright\arraybackslash} m{35em}}\toprule
\multicolumn{1}{c}{Symbol} & \multicolumn{1}{c}{Definition}
\\[1pt] \hline
 \,\,\, $I_t$\,\, & net inventory before replenishment in period $t$ \\[1pt] 
 \,\,\, $D_t$\,\, & demand in period $t$ \\[1pt] 
 \,\,\, $\phi$\,\, & demand distribution \\[1pt] 
 \,\,\, $b$\,\, & backlogging cost \\[1pt] 
 \,\,\, $h$\,\, & holding cost \\[1pt]  
  \,\,\, $T$\,\, & number of periods \\[1pt] 
  \,\,\, $q^{\rm r}_t$\,\, & regular order placed in period $t$ \\[1pt]  
  \,\,\, $q^{\rm e}_t$\,\, & regular order placed in period $t$ \\[1pt] 
 \,\,\, $c_{\rm r}$\,\, & cost of regular order \\[1pt] 
 \,\,\, $c_{\rm e}$\,\, & cost of expedited order \\[1pt] 
 \,\,\, $c_t$\,\, & total cost in period $t$ (\ie, sum of backlogging, holding, and order costs) \\[1pt] 
 \,\,\, $l_{\rm r}$\,\, & lead time of regular supplier \\[1pt] 
 \,\,\, $l_{\rm e}$\,\, & lead time of expedited supplier \\[1pt]
 \,\,\, $l$\,\, & difference between lead times $l_{\rm r}$ and $l_{\rm e}$ \\[1pt]
 \,\,\, $\mathbf{Q}^{\rm r}_{t}$\,\, & outstanding regular orders in period $t$ [\ie, $(q^{\rm r}_{t-l_{\rm r}},\dots,q^{\rm r}_{t-1})$] \\[1pt] 
 \,\,\, $\mathbf{Q}^{\rm e}_{t}$\,\, & outstanding expedited orders in period $t$ [\ie, $(q^{\rm e}_{t-l_{\rm e}},\dots,q^{\rm e}_{t-1})$] \\[1pt] 
 \,\,\, $ \mathbf{s}_t$\,\, & system state in period $t$ [\ie, $(I_{t},\mathbf{Q}_{t}^{\rm r},\mathbf{Q}_{t}^{\rm e})$] \\[1pt]  
 \,\,\,$\hat{\mathbf{a}}_t$\,\, & neural network actions in period $t$ [\ie, $(\hat{q}_t^{\rm r},\hat{q}_t^{\rm e})$] \\[1pt]   
 \,\,\,$\mathbf{w}$\,\, & neural network parameters \\[1pt] \bottomrule
\end{tabular}
\vspace{1mm}}
{An overview of the main variables used in the dual-sourcing model and NNCs.}
\end{table}
In the following sections, we provide an overview of different dual-sourcing heuristics that we use as baselines to evaluate the performance of neural network controllers (NNCs). Implementations of these sourcing policies and NNCs are available at \url{https://gitlab.com/ComputationalScience/inventory-optimization}. In Table~\ref{tab:model_variables}, we summarize the main variables used in the dual-sourcing model and NNCs.

Although we focus on a comparison between NNCs and capped dual index (CDI) policies in the main text, we will discuss related single-index and dual-index policies for the sake of completeness. In addition to single and dual-sourcing policies, our code base contains an implementation of the tailored base-surge (TBS) policy that can be used for further comparisons.
\section{Single Index}
\label{sec:single_index}
In accordance with \cite{scheller2007effective}, let $z_e$ and $z_r$ be the expedited and regular target order levels. The difference between $z_r$ and $z_e$ is denoted $\Delta$. For a single-index policy, expedited and regular orders are chosen such that the inventory position in period $t$ is $\tilde{I}_t=z_r-D_{t-1}=z_e+\Delta-D_{t-1}$. At the beginning of period $t$, an expedited order is placed according to
\begin{equation}
q_t^{\rm e}= [z_{\rm e}-\tilde{I}_t]^+=[D_{t-1}-\Delta]^+\,.
\label{eq:expedited_si}
\end{equation}
The regular order in period $t$ is
\begin{equation}
q_t^{\rm r}=[z_{\rm r}-(\tilde{I}_t+q_t^{\rm e})]^+=[D_{t-1}-[D_{t-1}-\Delta]^+]^+=D_{t-1}-q_t^{\rm e}=\min(\Delta,D_{t-1})\,.
\label{eq:regular_si}
\end{equation}
For the single index policy, the initial inventory level is $I_0=z_{\rm r}$ and for $t>l_{\rm r}$ the inventory evolution satisfies
\begin{align}
\begin{split}
I_{t+1}&=I_{t}+q_{t-l_{\rm r}}^{\rm r}+q_{t-l_{\rm e}}^{\rm e}-D_{t}=I_t+q_{t-l_{\rm e}}^{\rm e}-q_{t-l_{\rm r}}^{\rm e}-D_{t}+D_{t-l_{\rm r}-1}\\
&=I_{t-1}+q_{t-l_{\rm e}}^{\rm e}-q_{t-l_{\rm r}}^{\rm e}+q_{t-l_{\rm e}-1}^{\rm e}-q_{t-l_{\rm r}-1}^{\rm e}-D_{t}-D_{t-1}+D_{t-l_{\rm r}-1}+D_{t-l_{\rm r}-2}\\
&=z_{\rm r}+\sum_{i=0}^t\left( q_{t-l_{\rm e}-i}^{\rm e}-q_{t-l_{\rm r}-i}^{\rm e}\right)+\sum_{i=0}^t\left( D_{t-l_{\rm r}-1-i}-D_i\right)\\
&=z_{\rm r}+\sum_{i=t-l_{\rm r}+1}^{t-l_{\rm e}} q_i^{\rm e}-\sum_{i=t-l_{\rm r}}^t D_{i}\\
&=z_{\rm r}+\sum_{i=t-l_{\rm r}}^{t-l_{\rm e}-1} [D_i-\Delta]^+-\sum_{i=t-l_{\rm r}}^{t-l_{\rm e}-1} D_{i}-\sum_{i=t-l_{\rm e}}^{t} D_{i}\\
&=z_{\rm r}-\sum_{i=t-l_{\rm r}}^{t-l_{\rm e}-1} \min(\Delta,D_{i})-\sum_{i=t-l_{\rm e}}^t D_{i}\\
&=z_{\rm r}-d_1(\Delta)\,,
\end{split}
\label{eq:order_evolution_si}
\end{align}
where
\begin{equation}
d_1(\Delta)=\sum_{i=0}^{l_{\rm r}-l_{\rm e}-1} \min(\Delta,D_{i})+\sum_{i=0}^{l_{\rm e}} D_{i}\,.
\label{eq:D_Delta}
\end{equation}
We use $z_{\rm r}^*$ to denote the value of the target order level leading to a minimum cost. To find the optimal value of $z_{\rm r}^*$ , observe that the ordering costs are independent of $z_{\rm r}$ since they are proportional to Eqs.~\eqref{eq:expedited_si} and \eqref{eq:regular_si}. According to Eq.~\eqref{eq:order_evolution_si}, the optimal $z_{\rm r}^*$ can be determined by minimizing the expected holding and shortage costs
\begin{equation}
\begin{split}
&h \int_0^{z_{\rm r}} (z_{\rm r}-x)f_{d_1(\Delta)}(x)\,\mathrm{d}x+b\int_{d_1(\Delta)}^{\infty} (x-z_{\rm r}) f_{d_1(\Delta)}(x)\,\mathrm{d}x\\
&=h z_{\rm r} F_{d_1(\Delta)}(z_{\rm r})-h\int_0^{z_{\rm r}} x f_{d_1(\Delta)}(x)\,\mathrm{d}x-b z_{\rm r} [1-F_{d_1(\Delta)}(z_{\rm r})]+b\int_{d_1(\Delta)}^{\infty} xf_{d_1(\Delta)}(x)\,\mathrm{d}x\,,
\end{split}
\label{eq:critical_fractile}
\end{equation}
where $F_{d_1(\Delta)}(x)= \Pr(d_1(\Delta)\leq x)$ and $f_{d_1(\Delta)}(x)$ is the probability density function of $d_1(\Delta)$. Taking the derivative of Eq.~\eqref{eq:critical_fractile} with respect to $z_{\rm r}$ and evaluating at $z_{\rm r}=z_{\rm r}^*$ yields
\begin{equation}
h F_{d_1(\Delta)}(z_{\rm r}^*)-b[1-F_{d_1(\Delta)}(z_{\rm r}^*)]=0\,.
\end{equation}
The optimal regular target order level is thus given by the critical fractile
\begin{equation}
z_{\rm r}^*(\Delta) = F_{d_1(\Delta)}^{-1}\left(\frac{b}{b+h}\right)\,.
\label{eq:optimal_zr_si}
\end{equation}
The optimal values $(z_{\rm r}^*,\Delta^*)$ of the single index policy are found via the following optimization procedure.
\begin{enumerate}
    \item Iterate over all values of $\Delta \in [0,\dots,D_{\rm max}]$ (or use some search method such as a bisection method), where $D_{\rm max}$ is the maximum demand.
    \item For each $\Delta$, determine the distribution $F_{d_1(\Delta)}$ for a given $\Delta$ based on samples of $D(\Delta)$ that are calculated using Eq.~\eqref{eq:D_Delta}.
    \item For each $\Delta$, calculate $z_{\rm r}^*(\Delta)$ according to Eq.~\eqref{eq:optimal_zr_si}.
    \item Determine the $(z_{\rm r}^*,\Delta^*)$ that are associated with the smallest total cost.
\end{enumerate}
\section{Dual Index}
In the dual-index setting~\citep{veeraraghavan2008now}, one keeps track of the regular and expedited inventory positions
\begin{align}
\tilde{I}_{t+1}^{\rm e}&=\tilde{I}_{t}^{\rm e}+q_t^{\rm e}+q_{t-l}^{\rm r}-D_t=z_{\rm e}+O_t-D_t\,,\\
\tilde{I}_{t+1}^{\rm r}&=\tilde{I}_{t}^{\rm r}+q_t^{\rm e}+q_{t}^{\rm r}-D_t\,,
\end{align}
where $l=l_{\rm r}-l_{\rm e}>0$ and $O_t$ denotes the expedited inventory position overshoot. The dual-index expedited and regular orders are
\begin{align}
q_t^{\rm e}&=[z_{\rm e}-\tilde{I}_t^{\rm e}-q_{t-l}^{\rm r}]^+\,,\\
q_t^{\rm r}&=z_r-[\tilde{I}_{t}^{\rm r}+q_t^{\rm e}]^+=D_{t-1}-q_t^{\rm e}\label{eq:regular_di}\,,
\end{align}
respectively. Similar to the single-index strategy, the inventory evolution can be expressed in terms of target order levels and their difference $\Delta$:
\begin{equation}
I_{t+1}=z_{\rm e}+O_{t-l_{\rm e}}-\sum_{i=0}^{l_{\rm e}} D_{t-i}=z_{\rm e}-d_2(\Delta)\,,
\end{equation}
where $d_2(\Delta)=\sum_{i=0}^{l_{\rm e}} D_{t-i}-O_{t-l_{\rm e}}(\Delta)$. Let $G_{d_2(\Delta)}(x)= \Pr(d_2(\Delta)\leq x)$ denote the cumulative distribution function of $d_2(\Delta)$. In accordance with Eq.~\eqref{eq:optimal_zr_si}, one determines an optimal expedited target order level according to
\begin{equation}
z_{\rm e}^*(\Delta) = G_{d_2(\Delta)}^{-1}\left(\frac{b}{b+h}\right)\,.
\label{eq:optimal_zr_di}
\end{equation}
The optimal pair $(z_{\rm e}^*,\Delta^*)$ can be found using an iterative search similar to that described in Section~\ref{sec:single_index}. Note that unlike in the single-index policy, the quantity $\Delta^*$ may be larger than the maximum demand $D_{\rm max}$. 
\section{Capped Dual Index}
\label{sec:cdi}
The capped dual index policy \citep{sun2019robust} is an extension of the dual index policy and uses the following regular and expedited orders in period $t$:
\begin{equation}
q_t^{\rm r}=\min\left\{[{S_t^{\rm r *}}-I_t^{t+l-1}]^+,{\bar{q}_t^{\rm r *}}\right\}
\label{eq:qtr_cdi}
\end{equation}
and
\begin{equation}
q_t^{\rm e}=[{S_t^{\rm e *}}-I_t^t]^+\,.
\label{eq:qte_cdi}
\end{equation}
Here, we assume without loss of generality that $l_{\rm e}=0$. The quantity $I_t^{t+k}$ in Eqs.~\eqref{eq:qtr_cdi} and \eqref{eq:qte_cdi} denotes the sum of the net inventory level at the beginning of period $t$ and all in-transit orders that will arrive by period $t+k$. That is,
\begin{equation}
I_t^{t+k} = I_{t-1}+\sum_{i=t}^{\min(t+k,t-1)}q_i^{\rm e}+\sum_{i=t-l_{\rm r}}^{t-l_{\rm r}+k}q_i^{\rm r}\,,
\end{equation}
where $k=0,\dots,l_{\rm r}-1$. In accordance with \cite{sun2019robust}, we use the convention that $\sum_{i=a}^b=0$ if $a>b$. The parameters $(S_t^{\rm r *},{S_t^{\rm e *}},{\bar{q}_t^{\rm r *}})$ are found via a search procedure. If the demand distribution is time-independent, the CDI parameters are $S_t^{\rm r *}\equiv S^{\rm r *}$, ${S_t^{\rm e *}}\equiv {S^{\rm e *}}$, and ${\bar{q}_t^{\rm r *}}\equiv {\bar{q}^{\rm r *}}$.

Without using any search algorithm, the CDI parameters can be estimated according to
\begin{equation}
{S_t^{\rm e *}}=\frac{h\underline{D}_t^t+b \overline{D}_t^t}{h+b}\,, {S_t^{\rm r *}}=\frac{h\underline{D}_t^{t+l}+b \overline{D}_t^{t+l}}{h+b}
\end{equation}
and
\begin{equation}
{\bar{q}_t^{\rm r *}}=\frac{h(\underline{D}_t^{t+l}-\underline{D}_t^{t+l-1})+b (\overline{D}_t^{t+l}-\overline{D}_t^{t+l-1})}{h+b}\,,
\end{equation}
where $\underline{D}_t^n$ and $\overline{D}_t^n$ denote the minimum and maximum cumulative demand from period $t$ to $n$, respectively \citep{sun2019robust}.
\section{Optimal Policy and Value Iteration}
Finding an optimal policy that is associated with minimizing the expected cost per period $J$ [see Eq.~\eqref{eq:average_exp_cost}] can be obtained via the Bellman equation~\citep{bellman1954theory}. For a given arbitrary terminal cost function $v_0(\mathbf{s})$ the update

\begin{equation}
J_{t+1}(\mathbf{s})=\min_{\mathbf{a}_t\in \mathcal{A}_t}\left\{c_t(\mathbf{s}_t,\mathbf{a}_t)+\gamma \sum_{\mathbf{s}'\in\mathcal{S}_{t}} \Pr(\mathbf{s}_{t+1}=\mathbf{s}|\mathbf{s}'_t,\mathbf{a}_t)J_{t}(\mathbf{s}')\right\},\,  \mathbf{s}\in\mathcal{S}.
\label{eq:bellman}
\end{equation}

guarantees that $J^*=\lim\limits_{t\rightarrow \infty}\frac{J_t(\mathbf{s})}{t}$ \citep{bertsekas2011dynamic}. In practice, Eq.~\eqref{eq:bellman} can be used to solve optimization problems with small state space, and it has been used in the dual-sourcing literature as a benchmark of existing methods for small-scale instances \citep{scheller2007effective}.

Before describing the value iteration implementation, we state two simplifications that can be done without loss of generality. First, the lead time and the unit ordering cost of the expedited supplier can be set to zero: $l_{\rm e}=c_{\rm r}=0$ \citep{sheopuri2010new, sun2019robust}. Second, the dimensionality of the state space can be compressed to $l$ components \citep{sheopuri2010new}. We can define the expedited inventory position $\tilde{I}^{\rm e}_t=I_t+q_{t-l}^{\rm r}$ and compress the state vector to $\mathbf{s}_t = (\tilde{I}^{\rm e}_t, q^{\rm r}_{t-l+1}, \dots, q^{\rm r}_{t-1})$. 

Once actions ($q^{\rm r}_t, q^{\rm e}_t$) have been taken and demand $D_t$ is realized, the period cost is calculated as $f(\tilde{I}^{\rm e}_t+q^{\rm e}_t-D_t)$, where $f(x)=hx^++b[-x]^+$.
The state update equations become

\begin{equation}
\label{eq:compressed_state_update}
    \begin{cases}
    \tilde{I}^{\rm e}_{t+1} \leftarrow \tilde{I}^{\rm e}_{t} + q^{\rm e}_t + q^{\rm r}_{t-l+1} - D_t \\
    q^{\rm r}_{t-l+1} \leftarrow q^{\rm r}_{t-l+2}\\
    \dots\\
    q^{\rm r}_{t-2} \leftarrow q^{\rm r}_{t-1}\\
    q^{\rm r}_{t-1} \leftarrow q^{\rm r}_{t}
    \end{cases}
\end{equation}

For convenience, we define $\mathbf{Q} = (q^{\rm r}, q^{\rm e})$ and $\mathcal{D}_{\mathbf{Q}}$ the domain of optimal actions. The value iteration algorithm then proceeds as follows.
\begin{itemize}
    \item For each state $\mathbf{s} \in \mathcal{S}$, select an arbitrary initial cost $J_0(\mathbf{s})$
    \item For a given state $\mathbf{s}$ and action $\mathbf{Q}$, find the transition probabilities to state $\mathbf{s}'$ according to the demand distribution $\phi$. Let us denote those probabilities by $P(\mathbf{s}' | \mathbf{s}, \mathbf{Q})$. Calculate the cost $f(\mathbf{s}')$ associated with each transition $\mathbf{s}\xrightarrow{\mathbf{Q}} \mathbf{s}'$. Iterate those calculations for all combinations ${(\mathbf{s}, \mathbf{Q}) \in \mathcal{S}\times  \mathcal{D}_{\mathbf{Q}}}$.
    \item Apply the update $J_{k+1}(\mathbf{s}) = \min\limits_{\mathbf{Q} \in \mathcal{D}_{\mathbf{Q}}} \left\{ c_{\rm e}q^{\rm e} + 
    \sum\limits_{\mathbf{s}' \in \mathcal{S}} P(\mathbf{s}' | \mathbf{s}, \mathbf{Q})(f(\mathbf{s}')+J_{k+1}(\mathbf{s})) \right\}$, for all $\mathbf{s} \in \mathcal{S}$
    \item Calculate the expected cost approximation $\lambda_{k+1}(\mathbf{s}) = J_{k+1}(\mathbf{s}) / (k+1)$, for all $\mathbf{s} \in \mathcal{S}$
    \item Iterate the above update until $\max\limits_{\mathbf{s}\in\mathcal{S}}\left\{\lambda_{k+1}(\mathbf{s})-\lambda_{k}(\mathbf{s})\right\} < \epsilon$ 
\end{itemize}
\section{Neural-Network Structure and Learning Characteristics}
\label{sec:nn_structure_learning}
To provide insights into the representational capacity and learning characteristics of NNCs, we briefly discuss properties of (i) network structure as the basis of the potential representational power of a neural network, and (ii) learning dynamics and optimizers that help to learn effective inventory management policies. In the first part of this e-companion, we focus on the learning methods that we used in the dual-sourcing problems with synthetic demand (see Sections~\ref{sec:NNC_performance}--\ref{sec:fixed_costs}). The second part then summarizes implementation attributes associated with empirical demand example (see Section~\ref{sec:emp_demand}).
\subsection{Synthetic Demand}
\label{sec:synthetic_demand}
The network structure that we employ in dual-sourcing problems with synthetic demand distributions $\mathcal{U}\{0,4\}$ and $\mathcal{U}\{0,8\}$ (see Sections~\ref{sec:NNC_performance}--\ref{sec:fixed_costs}) uses the 
${(l_{\rm r}+l_{\rm e}+1)}$-dimensional state ${\mathbf{s}_t=(I_{t},\mathbf{Q}_{t}^{\rm r},\mathbf{Q}_{t}^{\rm e})}$ as an input and outputs actions $\hat{\mathbf{a}}_t=(\hat{q}_t^{\rm r},\hat{q}_t^{\rm e})$. We use fully connected layers with CELU activations, resembling the $\max(\cdot)$ operations that appear in the mathematical description of inventory management heuristics (see Sec.~\ref{sec:dual_sourcing_heuristics}). CELU activations \eqref{eq:celu} approximate ReLU activations in the limit $\alpha\rightarrow0$. In all numerical experiments, we use seven hidden layers with 128, 64, 32, 16, 8, 4, and 2 neurons, respectively. We set $\alpha=1$. Initial weights and biases are uniformly distributed according to $\mathcal{U}(-H_{\rm in}^{-1/2},H_{\rm in}^{-1/2})$, where $H_{\rm in}$ is the number of input features of the corresponding layer.

\cite{hanin2017approximating} and \cite{park2020minimum} formulated a universal approximation theorem and width bounds for neural networks with ReLU activations. In particular, they derived the conditions under which neural networks with ReLU activations can approximate any continuous, real-valued function arbitrarily well. These theorems supported our choice to use a neural network with CELU activations. Finally, (C)ELU activations are useful in many learning tasks because unlike ReLU functions they do not have vanishing gradients and mean activations near zero~\citep{clevert2015fast}.

We train neural networks using inventory time series with $T=100$ ($l_{\rm r}=2$), $150$ ($l_{\rm r}=3$), and $200$ ($l_{\rm r}=4$) periods and minibatches of size $512$. Large minibatch sizes and sufficiently long time series help to appropriately sample the action space. The use of large enough minibatches also aligns gradient descent in the direction of the global optimum~\citep{li2014efficient,masters2018revisiting}. 

We optimize the neural network weights $\mathbf{w}$ using RMSprop~\citep{rmsprob}, an adaptive learning rate method that is well-suited to perform mini-batch weights updates~\citep{kingma2014adam}. RMSprop uses the following update rules:
\begin{align}
\begin{split}
\mathbf{g}^{(n)}&=\nabla_{\mathbf{w}^{(n)}}   J^{(\hat{\pi}_t)}\,,\\
v^{(n+1)}&=\alpha v^{(n)} + (1-\alpha) [{\mathbf{g}^{(n)}}]^2\,,\\
\mathbf{w}^{(n+1)}&=\mathbf{w}^{(n)}-\frac{\eta}{\sqrt{v^{(n)}}+\epsilon} \mathbf{g}^{(n)} \,,
\end{split}
\end{align}
where $n$ denotes the current training epoch, $\eta$ is the learning rate, $\alpha$ is a smoothing constant, and $v$ is the weighted moving average of the squared gradient. The variable $\epsilon$ is used to improve numerical stability of the gradient-descent weight updates. In our numerical experiments, we set $\alpha=0.99$ and $\epsilon=10^{-8}$. Initially, the moving average of the squared gradient is $v^{(0)}=0$.

We primarily use a learning rate of $3\times 10^{-3}$ for the neural-network parameters and a learning rate of $10^{-1}$ for the adjustment of the initial net inventory. Adjusting the initial inventory during the learning process is helpful for NNCs to identify efficient policies on finite time horizons. As NNC policies approach optimal policies, it may help to lower the learning rates by up to an order of magnitude.

As detailed in the main text, standard backpropagation methods rely on real-valued gradients and neural-network parameters. To generate discrete actions (\ie, order quantities in inventory management problems), we detach the fractional part of an action from the underlying computational graph before generating the corresponding output (see Section~\ref{sec:frac_decoupling}).

\begin{figure}
    \FIGURE
    {\includegraphics{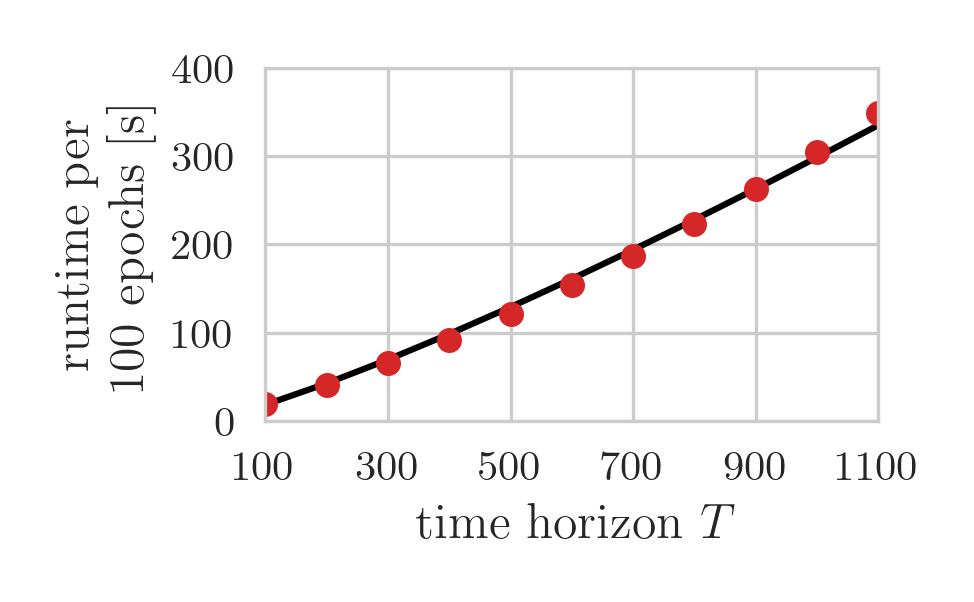}}
    {Training runtime as a function of time horizon $T$.\label{fig:runtime}}
    {Red disks show the runtime per 100 training epochs in seconds as a function of the time horizon of dual-sourcing dynamics for $h=5$, $b=495$, $l_{\rm r}=2$, and $l_{\rm e}=0$. The minibatch size is $M=512$. The solid black line is a guide-to-the-eye that is based on a power law with exponent $1.21$. Numerical experiments have been carried out on an AMD\textsuperscript{\textregistered} Ryzen threadripper 3970.}
\end{figure}

All numerical experiments have been performed on one CPU core. There are different factors that impact training time. For example, small learning rates of less than $10^{-3}$ were associated with a slow convergence in the initial stages of training. We have primarily used a learning rate of $3\times 10^{-3}$ to achieve convergence in about 5--30 minutes in a previously untrained neural network (\ie, without transfer learning). With transfer learning, training takes less than a minute for the majority of simulated instances. Other factors that impact training time include minibatch size $M$, neural-network depth and width, and the simulated time horizon $T$.

Figure \ref{fig:runtime} shows the runtime per 100 training epochs as a function of $T$ for the neural-network architecture described above and for a minibatch size of $M=512$. The solid black line is a guide-to-the-eye that is based on a power law with exponent $1.21$, suggesting that the runtime increases almost linearly with $T$ in the shown interval. Runtimes and the maximum value of $T$ that is still reasonable in training depend on both the minibatch size and used computer architecture. There is a tradeoff between large values of $T$ and large minibatch sizes $M$ that can both help learn policies that generalize well. 
\subsection{Empirical Demand}
\label{sec:empirical_demand}
To train NNCs that control dual-sourcing dynamics with empirical demand data in Section~\ref{sec:emp_demand}, we use as inputs in the first layer of the neural network the mean $\mu_t$ and standard deviation $\sigma_t$ of the truncated normal distribution \eqref{eq:empirical_demand_distribution}, and the reduced state $(I_t+q^{\rm r}_{t-l_{\rm r}},\tilde{\mathbf{Q}}_t^{\rm r},\mathbf{Q}_t^{\rm e})$, where $\tilde{\mathbf{Q}}_t^{\rm r}=(q^{\rm r}_{t-l_{\rm r}+1},\dots,q^{\rm r}_{t-1})$. The input state is therefore (${l_{\rm r}+l_{\rm e}+2}$)-dimensional. The hidden layer consists of 3 linear layers with 8 CELU activations each. Building on ideas from computer vision~\citep{fei2006one}, we employ a ``one-shot learning'' protocol to pretrain the neural network on one specific demand realization. The learning rate was initially varied between $1\times 10^{-4}$ and $3\times 10^{-3}$, and then set to $2\times 10^{-4}$ after reaching an expected cost per period of $8\times 10^5$. In the second step, we trained the NNC on 4 samples to improve performance. Learning rates varied between $1\times 10^{-4}$ and $3\times10^{-3}$. As in Section~\ref{sec:synthetic_demand}, we use RMSprop with the same hyperparameters. The total training time is between 20 minutes and about 1 hour on an Intel\textsuperscript{\textregistered} Core\textsuperscript{\texttrademark} i7-10510U CPU @ 1.80GHz.
\section{Low Service Level Performance}
\label{sec:low_service_level}
\begin{table}[hbt!]
\TABLE
{Expected costs per period of CDI and NNC order policies for low service levels.\label{tab:dual_sourcing_costs_low_service}}
{\centering
\renewcommand*{\arraystretch}{1.2}
\begin{tabular}{ >{\centering\arraybackslash} m{4em} >{\centering\arraybackslash} m{4em} 
>{\centering\arraybackslash} m{4em} 
>{\centering\arraybackslash} m{4em} 
>{\centering\arraybackslash} m{4em} 
>{\centering\arraybackslash} m{4em} 
>{\centering\arraybackslash} m{4em}
>{\centering\arraybackslash} m{4em}
>{\centering\arraybackslash} m{4em}}\toprule
$l_{\rm r}$ & $c_{\rm e}$ & demand & CDI cost & NNC cost & DP cost & CDI RMSE & NNC RMSE
\\[1pt]\hline\hline
2 & 5 & $\mathcal{U}\{0,4\}$ & 39.99 & \textbf{39.54} & 39.45 &0.60 &\textbf{0.32} \\[1pt]
2 & 5 & $\mathcal{U}\{0,8\}$ & 71.15 & \textbf{71.13} & 71.01 &\textbf{0.31} & 0.41\\[1pt]
2 & 10 & $\mathcal{U}\{0,4\}$ & 45.16 & \textbf{43.96} & 43.98 &0.69 &\textbf{0.25} \\[1pt]
2 & 10 & $\mathcal{U}\{0,8\}$ & 81.22 & \textbf{80.47} & 80.55 &0.65  &\textbf{0.53} \\[1pt]
2 & 20 & $\mathcal{U}\{0,4\}$ & \textbf{49.30} & \textbf{49.30} & 49.33 &\textbf{0.24} &0.40 \\[1pt]
2 & 20 & $\mathcal{U}\{0,8\}$ & \textbf{90.88} & \textbf{90.88} & 90.96 &0.79  &\textbf{0.46} \\[1pt]\midrule
3 & 5 & $\mathcal{U}\{0,4\}$ & 39.68 & \textbf{39.67} & 39.48 &0.62  &\textbf{0.60} \\[1pt]
3 & 5 & $\mathcal{U}\{0,8\}$ & 71.45 & \textbf{71.32} & 71.20 &\textbf{0.74} &1.27 \\[1pt]
3 & 10 & $\mathcal{U}\{0,4\}$ & 45.14 & \textbf{44.60} & 44.58 &0.69  &\textbf{0.36} \\[1pt]
3 & 10 & $\mathcal{U}\{0,8\}$ & 81.68 & \textbf{81.51} & 81.39 &0.94  &\textbf{0.57} \\[1pt]
3 & 20 & $\mathcal{U}\{0,4\}$ & 51.80 & \textbf{51.01} & 50.89 &0.46  &\textbf{0.45} \\[1pt]
3 & 20 & $\mathcal{U}\{0,8\}$ & 94.63 & \textbf{93.98} & 93.69 &0.96 &\textbf{0.55} \\[1pt]\bottomrule
\end{tabular}
\vspace{1mm}}
{CDI and NNC costs are based on $500$ realizations, each consisting of a time horizon $T=1,000$. For comparison, we also show the corresponding optimal, infinite-horizon DP cost. The parameters $l_{\rm r}$, $c_{\rm e}$, and the demand distribution are as listed in the first three columns. The remaining parameters are held constant at $c_{\rm r} = 0$, $b=85$, and $h = 15$.}
\end{table}
\begin{table}[hbt!]
\TABLE
{Expected costs per period of NNC order policies for low service levels with fixed order cost.\label{tab:dual_sourcing_costs_low_service_fixed_cost}}
{\centering
\renewcommand*{\arraystretch}{1.2}
\begin{tabular}{ >{\centering\arraybackslash} m{4em} 
>{\centering\arraybackslash} m{4em} 
>{\centering\arraybackslash} m{4em} 
>{\centering\arraybackslash} m{4em}
>{\centering\arraybackslash} m{4em}
>{\centering\arraybackslash} m{4em}}\toprule
$l_{\rm r}$ & $c_{\rm e}$ & demand & NNC cost & DP cost
\\[1pt]\hline\hline
2 & 5 & $\mathcal{U}\{0,4\}$ & 48.82 & 47.36 \\[1pt]
2 & 5 & $\mathcal{U}\{0,8\}$ & 82.28 & 81.44 \\[1pt]
2 & 10 & $\mathcal{U}\{0,4\}$ & 52.22 & 51.49 \\[1pt]
2 & 10 & $\mathcal{U}\{0,8\}$ & 89.28 & 89.57 \\[1pt]
2 & 20 & $\mathcal{U}\{0,4\}$ & 55.98 &  56.00 \\[1pt]
2 & 20 & $\mathcal{U}\{0,8\}$ & 98.05 &  98.37 \\[1pt]\midrule
3 & 5 & $\mathcal{U}\{0,4\}$ & 48.94 & 47.37 \\[1pt]
3 & 5 & $\mathcal{U}\{0,8\}$ & 82.31 & 81.66 \\[1pt]
3 & 10 & $\mathcal{U}\{0,4\}$ & 52.67 &  52.36 \\[1pt]
3 & 10 & $\mathcal{U}\{0,8\}$ & 90.19 & 90.90 \\[1pt]
3 & 20 & $\mathcal{U}\{0,4\}$ & 57.91 &  58.21 \\[1pt]
3 & 20 & $\mathcal{U}\{0,8\}$ & 101.36 & 101.84 \\[1pt]\bottomrule
\end{tabular}
\vspace{1mm}}
{NNC costs are based on a time horizon of $T=10,000$. For comparison, we also show the corresponding optimal, infinite-horizon DP cost. The parameters $l_{\rm r}$, $c_{\rm e}$, and the demand distribution are as listed in the first three columns. The remaining parameters are held constant at $c_{\rm r} = 0$, $b=85$, $h = 15$, $f_{\rm e}=10$, and $f_{\rm r}=5$.}
\end{table}
To complement the results provided in the main text for service levels $b/(h+b)$ of 95 and 99\%, we compare the performance of NNCs and CDI for a service level of 85\% (see Table~\ref{tab:dual_sourcing_costs_low_service}). As in the main text, we find that NNCs either outperform CDI or achieve equal performance in terms of expected costs per period.

For dual-sourcing dynamics with both fixed order costs and a low service level of 85\%, we find that NNCs are still able to produce expected costs per period that are near-optimal (see Table~\ref{tab:dual_sourcing_costs_low_service_fixed_cost}).
\section{Empirical Demand Data}
\subsection{Alternative Inputs}
\label{sec:empirical_baseline2}
\begin{figure}
    \FIGURE
    {\includegraphics{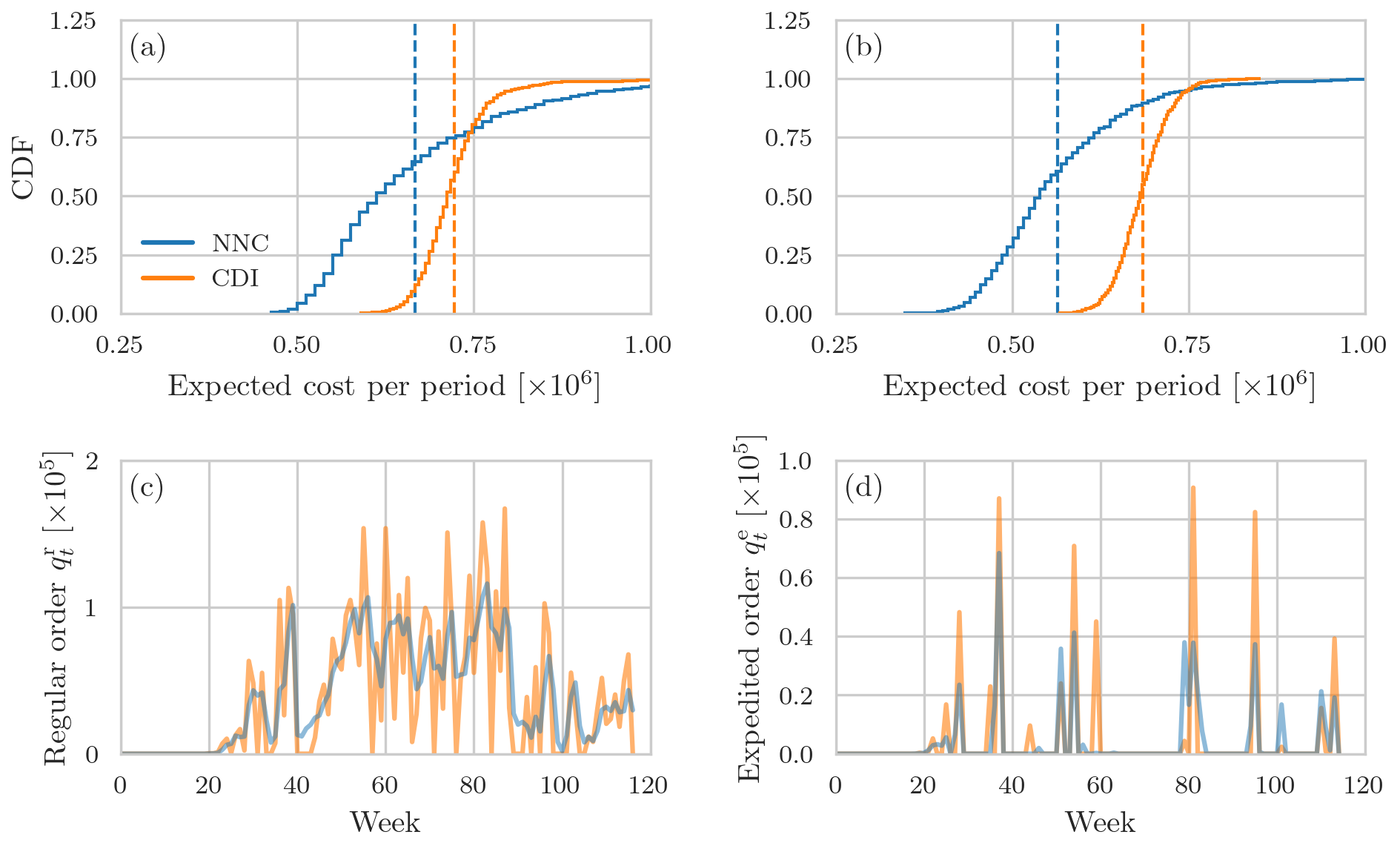}}
    {Controlling real-world inventory management problems using NNCs and CDI (alternative baseline).\label{fig:empirical_baseline2}}
    {(a,b) The cumulative distribution function of the average NNC and CDI costs. In panels (a) and (b), the backlog costs are $b=495$ and $b=95$, respectively. Shown results are based on $10^3$ i.d.d.\ samples. Dashed lines indicate mean expected costs per period [(a) 666,600 (NNC); 722,346 (CDI) and (b) 564,003 (NNC); 684,495 (CDI)]. (c) An example of regular orders generated by an NNC (solid blue line) and a CDI policy (solid orange line) for $b=495$. (d) An example of expedited orders generated by an NNC (solid blue line) and a CDI policy (solid orange line) for $b=495$. The remaining parameters of the underlying dual-sourcing problem are $h=5$, $c_{\rm r}=0$, $c_{\rm e}=20$, $l_{\rm r}=2$, and $l_{\rm e}=0$. The CDI baseline is based on Eq.~\eqref{eq:CDI_baseline2} (\ie, CDI has access to the Gaussian process moments $\mu_t,\sigma_t$ for times up to $t+l$). We use the same information for training NNCs.}
\end{figure}
In the application of NNCs to empirical demand data in Section~\ref{sec:emp_demand}, we use a CDI baseline where the potentially unknown Gaussian process moments for times $t'>t$ are estimated by $\mu_t$ and $\sigma_t$. Under the assumption that $\mu_t$ and $\sigma_t$ are known for times up to $t+l$, estimates of the CDI parameters ${S_t^{\rm r *}},{S_t^{\rm e *}},{\bar{q}_t^{\rm r *}}$ are provided by

\begin{align}
\begin{split}
{S_t^{\rm r *}}&=\frac{h \sum_{j=0}^l[\mu_{t+j}-2.58 \sigma_{t+j}]^++b \sum_{j=0}^l [\mu_{t+j}+2.58 \sigma_{t+j}]}{h+b}\\
{S_t^{\rm e *}}&=\frac{h [\mu_t-2.58 \sigma_t]^++b [\mu_t+2.58 \sigma_t]}{h+b}\\
{\bar{q}_t^{\rm r *}}&=\frac{h[\mu_{t+l}-2.58 \sigma_{t+l}]^+ + b[\mu_{t+l}+2.58 \sigma_{t+l}]}{h+b}\,,
\end{split}
\label{eq:CDI_baseline2}
\end{align}
where $l=l_{\rm r}-l_{\rm e}$ (see Section~\ref{sec:cdi}). Note that the estimate of ${S_t^{\rm e *}}$ in Eq.~\eqref{eq:CDI_baseline2} is the same as in Eq.~\eqref{eq:cdi_estimates} while the estimates of ${S_t^{\rm r *}}$ and ${\bar{q}_t^{\rm r *}}$ differ. In addition to the reduced state ${(I_t+q^{\rm r}_{t-1},\tilde{Q}_t^{\rm r},Q_t^{\rm e})}$ (see Section~\ref{sec:nn_structure_learning}), we also provide the neural network with the same quantities $[\mu_{t+j}-2.58 \sigma_{t+j}]^+,[\mu_{t+j}+2.58 \sigma_{t+j}]$ as in Eq.~\eqref{eq:CDI_baseline2} as inputs. The input state is therefore (${3l_{\rm r}-l_{\rm e}}$)-dimensional. The training process is as described in Section~\ref{sec:low_service_level}. To improve generalization performance of the learned policies, we used up to 512 samples after the initial ``one-shot'' training phase.

We generate 1,000 i.i.d.\ samples for $h=5$, $b=95,495$, $c_{\rm r}=0$, $c_{\rm e}=20$, $l_{\rm r}=2$, and $l_{\rm e}=0$. For $b=495$, the median expected costs per period are 620,968 (NNC) and 716,001 (CDI); the mean expected costs per period are 666,600 (NNC) and 722,346 (CDI) [dashed lines in Figure~\ref{fig:empirical_baseline2}(a)]. For $b=95$, the median expected costs per period are 541,150 (NNC) and 682,958 (CDI); the mean expected costs per period are 564,003 (NNC) and 684,495 (CDI) [dashed lines in Figure~\ref{fig:empirical_baseline2}(b)]. As in Section~\ref{sec:emp_demand} in the main text, we find that NNCs can significantly reduce the mean cost compared to a CDI policy. The relative reduction in mean cost is about 8\% and 18\% for $b=495$ and $95$, respectively.

Figure~\ref{fig:empirical_baseline2}(c,d) shows an example of regular and expedited orders that are generated by an NNC policy (solid blue lines) and a CDI policy (solid orange lines) for $b=495$. As in Figure~\ref{fig:empirical}(c) in the main text, NNC regular orders fluctuate to a lesser degree between very high and very low orders than those of CDI. The neural network also uses fewer expedited orders than CDI [see Figure~\ref{fig:empirical_baseline2}(d)].

To summarize, both methods achieve lower mean and median costs if information about the Gaussian process moments until times $t+l$ is used. Still, NNCs outperform CDI policies with substantial reductions in both mean and median cost. 
\subsection{Evaluation of LSTM and Transformer Architectures}
In this section, we complement the results associated with the NNC architecture from Section~\ref{sec:empirical_demand} that we use to control dual-sourcing dynamics with empirical demand data. We study the ability of long short-term memory (LSTM)~\citep{DBLP:journals/neco/HochreiterS97} and transformer architectures~\citep{vaswani2017attention,liu2021swin} to learn effective order policies without relying on inventory dynamics for recurrent connections as in the RNN from Section~\ref{sec:empirical_demand}.

The LSTM units that we implement are based on an input gate, output gate, and forget gate, which aim to memorize information across long time intervals.
Transformers use the attention mechanism to learn long-term dependencies in time-series data.
Although LSTM and transformer models have been successfully applied in different machine-learning tasks~\citep{DBLP:conf/nips/Bakker01,vaswani2017attention}, their training and hyperparameter optimization requires substantial computational resources to achieve state-of-the-art performance.
The following analysis can therefore only provide a starting point for the study of the application of such models to controlling inventory dynamics.

Our initial design assumption is that LSTMs and transformers are capable of learning long-term dependencies in a model-free way.
Therefore, we use information regarding the observed demand $D_{t}$ and initial inventory $I_0$ as inputs in both models.
Both architectures operate in an autoregressive manner. That is, the output of each recurrent layer at period $t$ is provided as input to the recurrent layer at period $t+1$. For transformers, we have to specify two inputs: (i) the encoder input that consists of all past timesteps, and (ii) the decoder input that consists of the sequence we plan to control. Unlike in other applications of transformers, here the decoder architecture needs to preserve causality in the decoder sequence~\citep{tay2020efficient}. To achieve this, we follow a common practice~\citep{tay2020efficient,vaswani2017attention} that is based on adding a decoder input mask where future periods are masked (\ie, input elements from future periods do not contribute to decoder calculations for the current period.)

Both LSTM and transformer architectures may suffer from exploding gradients and often require input normalization~\citep{henry2020query,laurent2016batch}.
To alleviate problems that result from outputs and losses with large values, we combine two methods. First, we use a stacked fully connected network to preprocess the inputs, and then provide the output of that network as an input to the LSTM/transformer layer.
Second, to further scale the output of the recurrent layer outputs, we also use a fully connected neural network as an output layer, which takes as an input the outputs of the recurrent layers.
\begin{table}
\TABLE
{Comparison of mean and median costs for dual-sourcing dynamics with empirical demand and $b=95$.\label{tab:b95}}
{\centering
\renewcommand*{\arraystretch}{1.2}
\begin{tabular}{llrr}\toprule
Moment inputs & Model & Mean cost & Median cost \\ 
\hline
\hline
\multirow[c]{3}{*}{current} 
 & CDI & 736,018 & 735,265 \\
 & NNC-LSTM & 853,517 & 657,263 \\
 & NNC-RNN & \bfseries 583,873 & \bfseries 563,711 \\
& NNC-Transformer & 765,312 & 599,672 \\ \hline
\multirow[c]{3}{*}{future} & CDI & 684,495 & 682,958 \\
 & NNC-LSTM & 742,677 & \bfseries 526,839 \\
 & NNC-RNN & \bfseries 564,003 & 541,150 \\
 & NNC-Transformer & 768,615 & 597,173 \\ \hline
\multirow[c]{2}{*}{none} 
 & NNC-LSTM & 839,096 & \bfseries 605,136 \\
 & NNC-Transformer & \bfseries 803,983 & 682,074 \\
 \bottomrule
\end{tabular}
\vspace{1mm}}
{We compare mean and median expected costs per period for CDI and NNCs that are based on the problem-tailored RNN (see main text), an LSTM, and a transformer. All results are based on $10^3$ i.i.d.\ samples. The parameters of the underlying dual-sourcing problem are $h=5$, $b=95$, $c_{\rm r}=0$, $c_{\rm e}=20$, $l_{\rm r}=2$, and $l_{\rm e}=0$. The row where moment inputs are ``current'' indicates that the corresponding results are based on models that have access to the Gaussian process moments $\mu_t,\sigma_t$ at period $t$, and ``future'' indicates that the models have access to the moments $\mu_t,\sigma_t$ for times up to $t+l$ (see Section~\ref{sec:empirical_baseline2}). LSTMs and transformers have been also trained without information on the underlying distribution moments (indicated by ``none'').}
\end{table}

By adding more layers, the neural network architecture becomes deeper and convergence deteriorates as vanishing gradients become more common. To address this problem, we introduce skip connections~\citep{ieeeskip} in LSTM architectures. Note that for the same reason deep transformer architectures also use this design approach by default~\citep{vaswani2017attention}.
To prepare the LSTM for controlling inventory dynamics, we pretrain the model on a supervised task where the sum of the output needs to approximate future demand. Then we continue training for cost minimization as in the RNN architecture that we use in the main text. For the transformer model, first train $10$ randomly initialized models and preserve the one with best validation loss. Then we further fine-tune the best performing model for $1,500$ epochs. We refer the reader to our code repository~\citep{GitLab} for more technical details on hyperparameters and training procedures.
\begin{table}
\TABLE
{Comparison of mean and median costs for dual-sourcing dynamics with empirical demand and $b=495$.\label{tab:b495}}
{\centering
\renewcommand*{\arraystretch}{1.2}
\begin{tabular}{llrr}\toprule
Moment inputs & Model & Mean cost & Median cost \\ 
\hline
\hline
\multirow[c]{3}{*}{current} 
 & CDI & 773,993 & 770,362 \\
 & NNC-LSTM & 1,227,035 & 1,045,389 \\
 & NNC-RNN & \bfseries 747,117 & \bfseries 692,118 \\ 
 & NNC-Transformer & 1,179,749 & 1,188,786 \\ \hline
\multirow[c]{3}{*}{future} 
 & CDI & 722,346 & 716,001 \\
 & NNC-LSTM & 1,188,504 & 1,103,443 \\
 & NNC-RNN & \bfseries 666,600 & \bfseries 620,968 \\
& NNC-Transformer & 1,300,017 & 1,242,873 \\ \hline
\multirow[c]{2}{*}{none} 
& NNC-LSTM & \bfseries 1,143,572 & \bfseries 1,012,889 \\
& NNC-Transformer & 1,206,373 & 1,239,072 \\
\bottomrule
\end{tabular}
\vspace{1mm}}
{We compare mean and median expected costs per period for CDI and NNCs that are based on the problem-tailored RNN (see main text), an LSTM, and a transformer. All results are based on $10^3$ i.i.d.\ samples. The parameters of the underlying dual-sourcing problem are $h=5$, $b=95$, $c_{\rm r}=0$, $c_{\rm e}=20$, $l_{\rm r}=2$, and $l_{\rm e}=0$. The row where moment inputs are ``current'' indicates that the corresponding results are based on models that have access to the Gaussian process moments $\mu_t,\sigma_t$ at period $t$, and ``future'' indicates that the models have access to the moments $\mu_t,\sigma_t$ for times up to $t+l$ (see Section~\ref{sec:empirical_baseline2}). LSTMs and transformers have been also trained without information on the underlying distribution moments (indicated by ``none'').}
\end{table}

The results for $b=95$ and $b=495$ are summarized in Tables~\ref{tab:b95} and \ref{tab:b495}, respectively. The remaining parameters of the underlying dual-sourcing problem are $h=5$, $c_{\rm r}=0$, $c_{\rm e}=20$, $l_{\rm r}=2$, and $l_{\rm e}=0$. The empirical demand distribution is as in Section~\ref{sec:empirical_demand}.
We observe that NNC-LSTMs and NNC-Transformers achieve good performance, even when no information on the underlying distribution moments is available. For $b=95$, the LSTM model achieves the best median cost of 526,839 when future mean and standard deviation are available (see Section~\ref{sec:empirical_baseline2}).
Still, the original NNC implementation achieves higher performance without extensive hyperparameter optimization as the use of inventory dynamics for defining recurrent connections introduces an inductive bias that seems to improve learning performance. 
Future work may study how the performance of LSTMs and transformers can be further optimized both in terms of longer training sessions and further hyperparameter optimization. 
Currently, a training session of NNC-LSTM takes about 2.5 minutes (including hyperparameter optimization). To find a well-performing model, we repeat this procedure 10 times and preserve the best model.
The transformer model requires approximately $15$ minutes of pre-training for $200$ epochs, where $10$ different initializations are tested.
Then the best performing model is preserved and it is fine-tuned for approximately $15$ minutes to achieve the reported performance.
Both architectures were deployed on a GPU, and the evaluation hardware is a laptop equipped with an NVIDIA RTX 3080 with 16GB of VRAM and an Intel i9 8-core CPU with 32 GB of RAM. When transformers are trained in a auto-regressive manner, their computational time performance is degraded, as the parallelization capabilities of the attention mechanism are not used efficiently. As another direction for future work, it would be interesting to combine LSTM or transformer architectures with the RNN approach from the main text. One possibility is to use the outputs of LSTM or transformer models as additional inputs to the (compressed) inventory state of the NNC-RNN controllers. 
\section{On the structure of solutions obtained by NNC}
In this section, we consider the actions obtained by NNC to study how they compare against known heuristics and the optimal solution. We first consider an instance with high relative backlog cost and high cost of expedited orders $(b=495, h=5, c_{\rm e}=20, l_{\rm r}=2, D_t\sim U\{0,4\})$. This instance has a two-dimensional state space, which we represent using the net inventory, $I_t$, and the inventory position, $I_t^{t+1}=I_t+q^{\rm r}_{t-1}$. Figure~\ref{fig:lr2} shows the ordering levels, obtained using NNC, of the regular order $q^{\rm r}_t$ (Panel \ref{fig:lr2qr}) and of the expedited order $q^{\rm e}_t$ (Panel \ref{fig:lr2qe}), respectively, for each state that belongs to the corresponding ergodic Markov chain.

\pgfplotsset{compat=1.16} 
\begin{figure}[htp]
\begin{subfigure}{0.49\textwidth}
  \centering    
    \begin{tikzpicture}[
         > = {Straight Barb[scale=1]},
dot/.style = {circle, draw, fill=#1, inner sep=2pt}
                     ] 
    \tikzset{arr/.style={latex-,shorten <= 2.5pt}}
\pgfplotstableread{tables/table_lr=2.dat}\lrtwotable
\begin{axis}[enlargelimits=false, 
                    xmin=1,
                    xmax=11,
                    xtick={1,...,11},
                    ymin=4,
                    ymax=11,
                    ytick={4,...,11},
            grid=both,
            separate axis lines,
            y axis line style= { draw opacity=0 },
            x axis line style= { draw opacity=0 }, thick]  
\addplot+[only marks, mark options={black, scale=1}] table[x=Itt, y=Itl] {\lrtwotable};
\addplot[->] coordinates {(1,4) (1, 11)} node[left, yshift=-1em, xshift=3.5em] {$I_t^{t+l_{\rm r}-1}$};
\addplot[->] coordinates {(1,4) (11,4)} node[left, yshift=1em] {$I_t$};
 \draw[fill=gray!80]
          (axis cs:2,5) -- (axis cs:3,5) -- (axis cs:4,6) -- (axis cs:6,6) -- (axis cs:3,5) -- (axis cs:2,5);
 \draw[fill=gray!60]
          (axis cs:3,6) -- (axis cs:4,7) -- (axis cs:5,8) -- (axis cs:8,8) -- (axis cs:7,7) -- (axis cs:4, 7);
 \draw[fill=gray!40]
          (axis cs:5.75,8.75) -- (axis cs:5.75,9.25) -- (axis cs:9.25,9.25) -- (axis cs:9.25,8.75) -- (axis cs:5.75,8.75);
 \draw[fill=gray!20]
          (axis cs:6.75,9.75) -- (axis cs:6.75,10.25) -- (axis cs:10.25,10.25) -- (axis cs:10.25,9.75) -- (axis cs:6.75,9.75); 
\draw[arr] (axis cs:5.75,9) -- (axis cs:4.75,9) node[left] {$q^{\rm r}_t=1$};
\draw[arr] (axis cs:6.75,10) --  (axis cs:5.75,10) node[left] {$q^{\rm r}_t=0$};
\draw[arr] (axis cs:5, 7.5) --  (axis cs:4, 7.5) node[left] {$q^{\rm r}_t=2$};
\draw[arr] (axis cs:4, 5.5) --  (axis cs:3, 5.5) node[left] {$q^{\rm r}_t=3$};
\end{axis}
\end{tikzpicture}
    \caption{$q^{\rm r}_t$ orders do not follow any known structure.}
    \label{fig:lr2qr}
\end{subfigure}%
\begin{subfigure}{0.49\textwidth}
  \centering    
    \begin{tikzpicture}[
         > = {Straight Barb[scale=1]},
dot/.style = {circle, draw, fill=#1, inner sep=2pt}
                     ] 
    \tikzset{arr/.style={latex-,shorten <= 2.5pt}}
\pgfplotstableread{tables/table_lr=2.dat}\lrtwotable
\begin{axis}[enlargelimits=false, 
                    xmin=1,
                    xmax=11,
                    xtick={1,...,11},
                    ymin=4,
                    ymax=11,
                    ytick={4,...,11},
            grid=both,
            separate axis lines,
            y axis line style= { draw opacity=0 },
            x axis line style= { draw opacity=0 }, thick]  
\addplot+[only marks, mark options={black, scale=1}] table[x=Itt, y=Itl] {\lrtwotable};
\addplot[->] coordinates {(1,4) (1, 11)} node[left, yshift=-1em, xshift=3.5em] {$I_t^{t+l_{\rm r}-1}$};
\addplot[->] coordinates {(1,4) (11,4)} node[left, yshift=1em] {$I_t$};
\draw[fill=gray!60]
    (axis cs:1.75,4.75) -- (axis cs:1.75,5.25) -- (axis cs:2.25,5.25) -- (axis cs:2.25,4.75) -- (axis cs:1.75,4.75);
\draw[fill=gray!40]
(axis cs:2.75,4.75) -- (axis cs:2.75,6.25) -- (axis cs:3.25,6.25) -- (axis cs:3.25,4.75) -- (axis cs:2.75,4.75);
\draw[fill=gray!20]
(axis cs:4,6) -- (axis cs:4,7) -- (axis cs:7,10) -- (axis cs:10,10) -- (axis cs:6,6) -- (axis cs:4,6);
\draw[arr] (axis cs:2,5.2) --  (axis cs:2,6.2) node[midway, yshift=1.3em] {$q^{\rm e}_t=2$};
\draw[arr] (axis cs:3,6.2) --  (axis cs:3,7.2) node[midway, yshift=1.3em] {$q^{\rm e}_t=1$};
\draw[arr] (axis cs:7,9.5) --  (axis cs:6,9.5) node[left] {$q^{\rm e}_t=0$};
\end{axis}
\end{tikzpicture}
    \caption{$q^{\rm e}_t$ orders have basestock structure.}
    \label{fig:lr2qe}
\end{subfigure}
\caption{Steady-state orders $(q^{\rm r}_t, q^{\rm e}_t)$ found by NNC ($l_{\rm r}=2, c_{\rm e}=20,b=495,h=5, D_t\in \mathcal{U}\{0, 4\}$).}
\label{fig:lr2}
\end{figure}

We observe that the net inventory remains always positive, between a minimum of 2 and a maximum of 10 units. The order values $q^{\rm r}_t$ exhibit higher sensitivity with respect to the inventory position as opposed to the net inventory, in the sense that for a given net inventory value higher than two units, there are multiple possible $q^{\rm r}_t$ values, depending on the inventory position. Fixing the inventory position to a certain value, however, results in a unique ordering quantity $q^{\rm r}_t$, with only exception being when the inventory position equals six units. Finally, it is worth noticing that the optimal $q^{\rm r}_t$ policy for this instance varies from NNC only in the state $(I_t, I_t^{t+1})=(2, 5)$, where it orders two units instead of three. The NNC policy, therefore, almost matches the optimal one, and it is also quite different from CDI: a CDI policy with a cap of 3 units and basestock level of 6 is the best one in this case. It ``merges'' points with inventory position of 7 with those of smaller inventory positions, resulting in suboptimal $q^{\rm r}_t$ actions for a total of six states. 

Orders from the expedited supplier, $q^{\rm e}_t$, have a basestock structure with respect to the net inventory, $I_t$. Specifically, we order from the expedited supplier to bring the net inventory to four units after replenishment, regardless of the inventory position. This is identical to the structure  of the optimal solution and the one obtained by CDI. It is worth noticing that the expedited order prevents the system from transiting to a negative net inventory state: backlog costs are 99 times higher than holding costs and about 25 times higher than expedited ordering, and as a result backlogs are avoided at all costs. Despite this, expedited orders are \textit{not} required 86\% of the time, which is the proportion of time the system spends in states where $q^{\rm e}_t=0$. In conclusion, whenever the net inventory and the inventory position have small levels it is optimal to order from both suppliers in order to avoid backlog costs (via expedited orders) and drive the system to higher net inventories in the future (via regular orders).

The second instance we consider has again $l_{\rm r}=2$ but smaller backlog ($b=95$) and expedited ordering ($c_{\rm e}=10$) costs. Figure \ref{fig:lr2b} compares a policy found by NNC (panel \ref{fig:lr2bqr}) to the optimal policy found by dynamic programming (panel \ref{fig:lr2bqe}). Note that each panel shows both $q_t^{\rm r}$ and $q_t^{\rm e}$. In contrast to the previous instance, we see that NNC has uncovered a more involved policy.

\begin{figure}[htb!]
\begin{subfigure}{0.49\textwidth}
  \centering    
    \begin{tikzpicture}[
         > = {Straight Barb[scale=1]},
dot/.style = {circle, draw, fill=#1, inner sep=2pt}
                     ] 
    \tikzset{arr/.style={latex-,shorten <= 2.5pt}}
\pgfplotstableread{tables/table_lr=2b.dat}\lrthreetable
\begin{axis}[enlargelimits=false, 
                    xmin=-1,
                    xmax=10,
                    xtick={0,...,10},
                    ymin=0,
                    ymax=10,
                    ytick={0,...,10},
            grid=both,
            separate axis lines,
            y axis line style= { draw opacity=0 },
            x axis line style= { draw opacity=0 }, thick]  
\addplot+[only marks, mark options={black, scale=1}] table[x=Itt, y=Itl] {\lrthreetable};
\addplot[->] coordinates {(0,0) (0, 10)} node[right, yshift=-1em] {$I_t^{t+l_{\rm r}-1}$};
\addplot[->] coordinates {(0,0) (10,0)} node[left, yshift=1em] {$I_t$};
 \draw[fill=gray!80]
          (axis cs:-0.25,2.25) -- (axis cs:0.25,2.25) -- (axis cs:0.25,1.75) -- (axis cs:-0.25,1.75) -- (axis cs:-0.25,2.25);
  \draw[fill=gray!60]
           (axis cs:1,3) -- (axis cs:2,4) -- (axis cs:2,5) -- (axis cs:4,7) -- (axis cs:5,7) --  (axis cs:5,6) --  (axis cs:6,6) --  (axis cs:5,6) --  (axis cs:5,5) --  (axis cs:4,4) --  (axis cs:2,4) -- (axis cs:2,5) -- (axis cs:2,4) --  (axis cs:1,3);
\draw[fill=gray!40]
            (axis cs:4.75, 8.25) -- (axis cs:6.25, 8.25) -- (axis cs:6.25, 7.25) -- (axis cs:7.25, 7.25) -- (axis cs:7.25, 7.75) -- (axis cs:7.75, 7.75) -- (axis cs:7.75, 8.25) -- (axis cs:8.25, 8.25) -- (axis cs:8.25, 7.5) -- (axis cs:7.75, 7.5) -- (axis cs:7.75, 6.75) -- (axis cs:5.75, 6.75) -- (axis cs:5.75, 7.75) -- (axis cs:4.75, 7.75) -- (axis cs:4.75, 8.25);
\draw[fill=gray!20]
            (axis cs: 5.75, 8.75) -- (axis cs: 5.75, 9.25) -- (axis cs: 9.25, 9.25) -- (axis cs: 9.25, 8.75) -- (axis cs: 7.25, 8.75) -- (axis cs: 7.25, 7.75) -- (axis cs: 6.75, 7.75) -- (axis cs: 6.75, 8.75) -- (axis cs: 5.75, 8.75);
\draw[thick, dashed, pattern=north east lines]
            (axis cs: -0.5, 1.5) -- (axis cs: -0.5, 2.5) -- (axis cs:0.5, 2.5) -- (axis cs: 0.5, 1.5) -- (axis cs: -0.5, 1.5);
\draw[thick, dashed, pattern=north east lines]
            (axis cs: 0.5, 2.5) -- (axis cs: 0.5, 3.5) -- (axis cs:1.5, 3.5) -- (axis cs: 1.5, 2.5) -- (axis cs: 0.5, 2.5);
\draw[thick, dashed, pattern=north east lines]
            (axis cs: 1.5, 3.5) -- (axis cs: 1.5, 4.5) -- (axis cs:2.5, 4.5) -- (axis cs: 2.5, 3.5) -- (axis cs: 1.5, 3.5);
\draw[thick, dashed, pattern=north east lines]
            (axis cs: 1.75, 5.25) -- (axis cs: 3.25, 5.25) -- (axis cs:3.25, 3.75) -- (axis cs: 2.75, 3.75) -- (axis cs: 2.75, 4.75) -- (axis cs: 2.75, 4.75) -- (axis cs: 1.75, 4.75) -- (axis cs: 1.75, 5.25);
\draw[arr] (axis cs:6.5,7.) -- (axis cs:7.5,6) node[right] {$q^{\rm r}_t=1$};
\draw[arr] (axis cs:6,9) --  (axis cs:5,9) node[left] {$q^{\rm r}_t=0$};
\draw[arr] (axis cs:4.5, 5.5) -- (axis cs:5.5, 4.5) node[right] {$q^{\rm r}_t=2$};
\draw[arr] (axis cs:0., 2.) -- (axis cs:2., 2.) node[right] {$q^{\rm r}_t=3$};
\draw[arr] (axis cs:0.5, 1.5) -- (axis cs:1.5, .5) node[right] {$q^{\rm e}_t=4$};
\draw[arr] (axis cs:1, 3.5) -- (axis cs:1., 5.5) node[midway, yshift=1.7em] {$q^{\rm e}_t=3$};
\draw[arr] (axis cs:2.5, 3.5) -- (axis cs:4., 3.) node[right] {$q^{\rm e}_t=2$};
\draw[arr] (axis cs:2.5, 5.2) -- (axis cs:2.5, 7.25) node[midway, yshift=1.7em] {$q^{\rm e}_t=1$};
\end{axis}
\end{tikzpicture}
    \caption{$(q^{\rm r}_t, q^{\rm e}_t)$ orders found by NNC Cost=$19.881$.}
    \label{fig:lr2bqr}
\end{subfigure}%
\begin{subfigure}{0.49\textwidth}
  \centering    
    \begin{tikzpicture}[
         > = {Straight Barb[scale=1]},
dot/.style = {circle, draw, fill=#1, inner sep=2pt}
                     ] 
    \tikzset{arr/.style={latex-,shorten <= 2.5pt}}
\pgfplotstableread{tables/table_lr=2b_dp.dat}\lrthreetable
\begin{axis}[enlargelimits=false, 
                    xmin=-1,
                    xmax=10,
                    xtick={0,...,10},
                    ymin=0,
                    ymax=10,
                    ytick={0,...,10},
            grid=both,
            separate axis lines,
            y axis line style= { draw opacity=0 },
            x axis line style= { draw opacity=0 }, thick]  
\addplot+[only marks, mark options={black, scale=1}] table[x=Itt, y=Itl] {\lrthreetable};
\addplot[->] coordinates {(0,0) (0, 10)} node[right, yshift=-1em] {$I_t^{t+l_{\rm r}-1}$};
\addplot[->] coordinates {(0,0) (10,0)} node[left, yshift=1em] {$I_t$};
\draw[fill=gray!60]
    (axis cs:1.,3) -- (axis cs:4,6) -- (axis cs:6,6) -- (axis cs:5,5) -- (axis cs:4,5) -- (axis cs:3,4) -- (axis cs:2,4) -- (axis cs:1,3);
\draw[fill=gray!40]
    (axis cs:5.,7) -- (axis cs:6,8) -- (axis cs:8,8) -- (axis cs:7,7) -- (axis cs:5,7);
 \draw[fill=gray!20]
    (axis cs:7.75,8.75) -- (axis cs:7.75,9.25) -- (axis cs:9.25,9.25) -- (axis cs:9.25,8.75) -- (axis cs:7.75,8.75);
\draw[arr] (axis cs:8, 9) -- (axis cs: 6.5,9) node[left]{$q^{\rm r}_t=0$};
\draw[arr] (axis cs:7, 7.5) -- (axis cs: 8,6.5) node[right]{$q^{\rm r}_t=1$};
\draw[arr] (axis cs:5, 5.5) -- (axis cs: 6,4.5) node[right]{$q^{\rm r}_t=2$};
\draw[dashed, pattern=north east lines]
            (axis cs: .5, 2.5) -- (axis cs: 0.5, 3.5) -- (axis cs:1.5, 3.5) -- (axis cs: 1.5, 2.5) -- (axis cs: 0.5, 2.5);
\draw[dashed, pattern=north east lines]
            (axis cs: 1.5, 3.5) -- (axis cs: 1.5, 4.5) -- (axis cs:2.4, 4.5) -- (axis cs: 2.4, 3.5) -- (axis cs: 1.5, 3.5);
\draw[dashed, pattern=north east lines]
            (axis cs: 2.6, 3.5) -- (axis cs: 2.6, 5.5) -- (axis cs:3.5, 5.5) -- (axis cs: 3.5, 3.5) -- (axis cs: 2.6, 3.5);
\draw[arr] (axis cs:1.,3.5) --  (axis cs:1.,5.5) node[midway, yshift=1.6em] {$q^{\rm e}_t=3$};
\draw[arr] (axis cs:2.,4.5) --  (axis cs:2.,6.5) node[midway, yshift=1.6em] {$q^{\rm e}_t=2$};
\draw[arr] (axis cs:3,5.5) --  (axis cs:3.,7.5) node[midway, yshift=1.6em] {$q^{\rm e}_t=1$};
\end{axis}
\end{tikzpicture}
    \caption{$(q^{\rm r}_t, q^{\rm e}_t)$ orders found by DP. Cost=$19.730$}
    \label{fig:lr2bqe}
\end{subfigure}
\caption{NNC and DP policies. Hashed areas show expedited orders. ($l_{\rm r}=2, c_{\rm e}=10,b=95,h=5, D_t\in \mathcal{U}\{0, 4\}$).}
\label{fig:lr2b}
\end{figure}

A careful examination of Figure \ref{fig:lr2b} reveals two facts. First, NNC leads to a larger state space compared to dynamic programming as there are 25 recurrent states instead of 17. Second, the policy NNC found is not a simple function of $I_t$ and $I_{t}^{t+l_{\rm r}-1}$. The optimal solution follows a base stock policy for the expedited supplier with a level of four units, while the policy for the regular supplier is similar to the one from the previous instance, namely a CDI-like policy with a variable cap. Although the state spaces and policies look quite different, NNC's policy has a $0.8\%$ gap compared to the optimal one, attaining an expected cost of 19.881 instead of 19.730. An important conclusion from this example is that although NNC policies can be close to optimal, there is no guarantee that they have an interpretable structure, even when the optimal solution itself has a relatively simple structure.

The final instance we consider has two key differences: first, the backlog cost is of same order of magnitude as the holding cost, and second, the lead time of the regular supplier is three periods instead of two $(b=60, h=40, c_{\rm e}=10, l_{\rm r}=3, D_t\sim U\{0,4\})$. In this system, backlogging and holding costs are both expensive relative to the expedited orders, and therefore there is incentive to use such orders to drive the system in a balance between these costs. Figure \ref{fig:lr3} reports on the results.
\begin{figure}[htb!]
\begin{subfigure}{0.49\textwidth}
  \centering    
    \begin{tikzpicture}[
         > = {Straight Barb[scale=1]},
dot/.style = {circle, draw, fill=#1, inner sep=2pt}
                     ] 
    \tikzset{arr/.style={latex-,shorten <= 2.5pt}}
\pgfplotstableread{tables/table_lr=3.dat}\lrthreetable
\begin{axis}[enlargelimits=false, 
                    xmin=-3,
                    xmax=7,
                    xtick={-3,...,7},
                    ymin=-1,
                    ymax=7,
                    ytick={-1,...,7},
            grid=both,
            separate axis lines,
            y axis line style= { draw opacity=0 },
            x axis line style= { draw opacity=0 }, thick]  
\addplot+[only marks, mark options={black, scale=1}] table[x=Itt, y=Itl] {\lrthreetable};
\addplot[->] coordinates {(0,-2) (0, 7)} node[left, yshift=-1em] {$I_t^{t+l_{\rm r}-1}$};
\addplot[->] coordinates {(-3,0) (7,0)} node[left, yshift=-1em] {$I_t$};
 \draw[fill=gray!50]
          (axis cs:3,5) -- (axis cs:5,5) -- (axis cs:2,2) -- (axis cs:0,2) -- (axis cs:0,1) -- (axis cs:-1,0) -- (axis cs:-2, 0) -- (axis cs:3, 5);
 \draw[fill=gray!10]
          (axis cs:3.75,6.25) -- (axis cs:6.25,6.25) -- (axis cs:6.25,5.75) -- (axis cs:3.75,5.75) --  (axis cs:3.75,6.25);
\draw[arr] (axis cs:2.5,3.5) -- (axis cs:4,3) node[right] {$q^{\rm r}_t=1$};
\draw[arr] (axis cs:3.8,6) --  (axis cs:2,6) node[left] {$q^{\rm r}_t=0$};
\end{axis}
\end{tikzpicture}
    \caption{$q^{\rm r}_t$ orders have capped basestock structure.}
    \label{fig:lr3qr}
\end{subfigure}%
\begin{subfigure}{0.49\textwidth}
  \centering    
    \begin{tikzpicture}[
         > = {Straight Barb[scale=1]},
dot/.style = {circle, draw, fill=#1, inner sep=2pt}
                     ] 
    \tikzset{arr/.style={latex-,shorten <= 2.5pt}}
\pgfplotstableread{tables/table_lr=3.dat}\lrthreetable
\begin{axis}[enlargelimits=false, 
                    xmin=-3,
                    xmax=7,
                    xtick={-3,...,7},
                    ymin=-1,
                    ymax=7,
                    ytick={-1,...,7},
            grid=both,
            separate axis lines,
            y axis line style= { draw opacity=0 },
            x axis line style= { draw opacity=0 }, thick]  
\addplot+[only marks, mark options={black, scale=1}] table[x=Itt, y=Itl] {\lrthreetable};
\addplot[->] coordinates {(0,-2) (0, 7)} node[left, yshift=-1em] {$I_t^{t+l_{\rm r}-1}$};
\addplot[->] coordinates {(-3,0) (7,0)} node[left, yshift=-1em] {$I_t$};
\draw[fill=gray!50]
    (axis cs:-2.25,-0.25) -- (axis cs:-1.75,-0.25) -- (axis cs:-1.75,0.25) -- (axis cs:-2.25,0.25) -- (axis cs:-2.25,-0.25);
\draw[fill=gray!40]
(axis cs:-1.25,-0.25) -- (axis cs:-1.25,1.25) -- (axis cs:-0.75,1.25) -- (axis cs:-0.75,-0.25) -- (axis cs:-1.25,-0.25);
\draw[fill=gray!30]
(axis cs:-0.25,0.75) -- (axis cs:-0.25,2.25) -- (axis cs:0.25,2.25) -- (axis cs:0.25,0.75) -- (axis cs:-0.25,0.75);
\draw[fill=gray!20]
(axis cs:0.75,1.75) -- (axis cs:0.75,3.25) -- (axis cs:1.25,3.25) -- (axis cs:1.25,1.75) -- (axis cs:0.75,1.75);
\draw[fill=gray!10]
(axis cs:2,4) -- (axis cs:4,6) -- (axis cs:6,6) -- (axis cs:2,2) -- (axis cs:2,4);
\draw[arr] (axis cs:2.5,3.5) -- (axis cs:4,3) node[right] {$q^{\rm e}_t=0$};
\draw[arr] (axis cs:1.,2.25) --  (axis cs:2.25,1.75) node[right] {$q^{\rm e}_t=1$};
\draw[arr] (axis cs:0.,1.25) --  (axis cs:1.25,0.75) node[right] {$q^{\rm e}_t=2$};
\draw[arr] (axis cs:-1,0.25) --  (axis cs:0.25,-0.33) node[right] {$q^{\rm e}_t=3$};
\draw[arr] (axis cs:-2,0.25) --  (axis cs:-2,1.25) node[midway, yshift=1.2em] {$q^{\rm e}_t=4$};
\end{axis}
\end{tikzpicture}
    \caption{$q^{\rm e}_t$ orders have basestock structure.}
    \label{fig:lr3qe}
\end{subfigure}
\caption{Steady-state orders $(q^{\rm r}_t, q^{\rm e}_t)$ found by NNC ($l_{\rm r}=3, c_{\rm e}=10,b=60,h=40, D_t\in \mathcal{U}\{0, 4\}$).}
\label{fig:lr3}
\end{figure}

Note that in this case the state space is three-dimensional, consisting of the net inventory and two inventory positions, namely $I_t^{t+1}$ and $I_t^{t+2}$. However, we observed that the ordering decisions depend only on the net inventory and the total inventory position $I_t^{t+2}$, and therefore we can project the state space in those two dimensions. In general, we could project the state space to those two dimensions by averaging over the projected states, but it was not necessary to do so for this instance. Similar to the results reported in Figure~\ref{fig:lr2}, expedited orders follow a base stock structure (panel \ref{fig:lr3qe}). This time, however, they are utilized more commonly, because the cost of driving the system to better states is lower than the long-run benefit resulting from being in those states. The system spends about 55\% of time in states where a positive expedited order is in place. Regular orders (panel \ref{fig:lr3qr}) follow a CDI policy with a basestock value of six and a cap of one unit, relative to the inventory position. The regular order here depends solely on whether the total inventory position is below six units or not, while the expedited order is sensitive to the current net inventory level. A regular order of one unit is ordered 99\% of the time, making this policy almost identical to the tailored base-surge (TBS) heuristic. Overall, these observations confirm earlier analytical results on the structure of optimal solutions, and in particular that $0\geq \frac{\partial q^{\rm r}_t}{\partial I_t^t}\geq\frac{\partial q^{\rm r}_t}{\partial I_t^{t+1}} \geq\dots\geq \frac{\partial q^{\rm r}_t}{\partial I_t^{t+l_{\rm r}-1}}\geq -1$ and $0\geq \frac{\partial q^{\rm e}_t}{\partial I_t^{t+l_{\rm r}-1}}\geq\frac{\partial q^{\rm e}_t}{\partial I_t^{t+l_{\rm r}-2}} \geq\dots\geq \frac{\partial q^{\rm e}_t}{\partial I_t^{t}}\geq -1$. Moreover, they suggest that if the optimal expedited orders, $q^{\rm e}_t$, have a basestock structure, the basestock value is influenced by the relative cost of $c_{\rm e}$ compared to driving the system to states that optimally balance backlogging and holding costs. Finally, although NNC policies appear to attain near-optimal solutions, and that they may resemble optimal policies, their structure can be rather arbitrary. Hence, it is challenging to utilize them to obtain analytical insights on the structure of near-optimal solutions. Future work may study NNC policies that are obtained using additional constraints that bias the learning towards certain policy structures. 
\ACKNOWLEDGMENT{LB acknowledges financial support from the Swiss National Fund (grant number P2EZP2\_191888) and the Army Research Office (grant number W911NF-23-1-0129). TA acknowledges that the research was supported by NCCR Automation, a National Centre of Competence in Research, funded by the Swiss National Science Foundation (grant number 180545).}

\end{document}